%% file: paper-5066.tex
\documentclass[runningheads]{llncs}
\usepackage{graphicx}

\usepackage{tikz}
\usepackage{comment}
\usepackage{amsmath,amssymb} 
\usepackage{color}
\usepackage{pifont}
\usepackage[accsupp]{axessibility}  
\usepackage{subcaption}
\usepackage[pagebackref,breaklinks,colorlinks]{hyperref}
\usepackage{threeparttable}

\usepackage[capitalize]{cleveref}

\usepackage[lined, ruled]{algorithm2e}

\usepackage{listings}
\usepackage{orcidlink}

\usepackage{xcolor}

\usepackage{booktabs}
\usepackage{multirow}
\usepackage{cite}

\newcommand{\ra}[1]{\renewcommand{\arraystretch}{#1}}

\newcommand{\xredit}[1]{#1}

\newcommand{\specialcell}[2][c]{\begin{tabular}[#1]{@{}c@{}}#2\end{tabular}}

\makeatletter
\DeclareRobustCommand\onedot{\futurelet\@let@token\@onedot}
\def\@onedot{\ifx\@let@token.\else.\null\fi\xspace}

\def\eg{\emph{e.g}\onedot} 
\def\ie{\emph{i.e}\onedot} 
 
\def\etc{\emph{etc}\onedot} 
\def\wrt{w.r.t\onedot} 
 
\def\etal{\emph{et al}\onedot}
\makeatother

\begin{document}
\pagestyle{headings}
\mainmatter
\def\ECCVSubNumber{5066}  

\title{Semi-Supervised Keypoint Detector and Descriptor for Retinal Image Matching} 

\titlerunning{SuperRetina for Retinal Image Matching}
%
\author{Jiazhen Liu\inst{1, 2}\orcidlink{0000-0003-0584-4571} \and
Xirong Li\inst{1, 2}\thanks{Corresponding author: Xirong Li (xirong@ruc.edu.cn)}\orcidlink{0000-0002-0220-8310} \and
Qijie Wei\inst{2,3} \and 
Jie Xu\inst{4} \and 
Dayong Ding \inst{3}}
\authorrunning{J. Liu et al.}
%
\institute{
MoE Key Lab of DEKE, Renmin University of China
\and
AIMC Lab, School of Information, Renmin University of China
\and
Vistel AI Lab, Visionary Intelligence Ltd, Beijing, China
\and 
Institute of Ophthalmology, Tongren Hospital, Beijing, China
}
\maketitle

\begin{abstract}
For retinal image matching (RIM), we propose \emph{SuperRetina}, the first end-to-end method with jointly trainable keypoint detector and descriptor. 
SuperRetina is trained in a novel semi-supervised manner. A small set of (nearly 100) images are incompletely labeled and used to supervise the network to detect keypoints on the vascular tree. To attack the incompleteness of manual labeling, we propose Progressive Keypoint Expansion to enrich the keypoint labels at each training epoch. By utilizing a keypoint-based improved triplet loss as its description loss, SuperRetina produces  highly discriminative descriptors at full input image size. Extensive experiments on multiple real-world datasets justify the viability of SuperRetina. Even with manual labeling replaced by auto labeling and thus making the training process fully manual-annotation free, SuperRetina compares favorably against a number of strong baselines for two RIM tasks, \ie image registration and identity verification. 

\end{abstract}

\section{Introduction}
\label{sec:intro}

\input{intro}

\section{Related Work}
\label{sec:related}
\input{related}

\section{Proposed Method}
\label{sec:method}
\input{method}

\section{Evaluation} \label{sec:eval}

\input{experiment}

\section{Conclusions} \label{sec:concs}

Real-world experiments allow us to conclude as follows. The proposed PKE strategy is effective for resolving the incompleteness of manual labeling for semi-supervised training, improving mAUC from 0.685 to 0.755 for retinal image registration on the FIRE dataset and reducing EER from 5.14$\%$ to 0.83$\%$ for retina-based identity verification on the most challenging CLINICAL dataset. SuperRetina beats the best baselines, \ie  REMPE for image registration (mAUC: 0.755 versus 0.720), and SuperPoint for  identity verification (EER: 0.83$\%$ versus 1.06$\%$ on CLINICAL, 1.18$\%$ versus 2.00$\%$ on BES). Even with the manually labeled training data fully replaced by auto-labeling, and thus making the training process fully manual annotation free, SuperRetina preserves mostly its performance and compares favorably against the previous methods for RIM.

\medskip

\textbf{Acknowledgments}. This work was supported by NSFC (No. 62172420, No. 62072463), BJNSF (No. 4202033), and Public Computing Cloud, Renmin University of China.

\clearpage

\bibliographystyle{splncs04}
\bibliography{egbib}

\clearpage
\input{supplementary/supplementary}

%
%

\end{document}

%% file: intro.tex
This paper is targeted at retinal image matching (RIM), which is to match color fundus photographs based on their visual content. Matching criteria are task dependent. As the retinal vasculature is known be unique, stable across ages and naturally anti-counterfeiting \cite{simon1935new}, retinal images are used for high-security \emph{identity verification} \cite{oinonen2010identity}. In this context, two retinal images are considered matched if they were taken from the same eye. RIM is also crucial for  \emph{retinal image registration}, which is to geometrically align two or more  images taken from different regions of the same retina (at different periods). Aligned images can be used for  wide-field imaging \cite{cattin2006retina}, precise cross-session assessment of retinal condition progress \cite{hernandez2020rempe}, and accurate laser treatment on the retina \cite{truong2019glampoints}. RIM is thus a valuable topic in computer vision.

Developing a generic method for RIM is nontrivial. Due to varied factors in fundus photography such as illumination condition, abnormal retinal changes and natural motions of the fixating eye, retinal images of the same eye may vary significantly in terms of their visual appearance. Common lesions in diabetic
retinopathy such as microaneurysm and intraretinal hemorrhage appear as dark dots, while cotton-wool spots look like white blobs \cite{icpr20-lesion-net}. The classical SIFT detector \cite{lowe2004distinctive}, which finds corners and blobs in a scale-invariant manner, tends to respond around the lesions and the boundary between the circular foreground and the dark background, see \cref{fig:matching}. SIFT keypoints detected at these areas lack both repeatability and reliability.

Recently, GLAMpoints \cite{truong2019glampoints} is proposed as a trainable detector for RIM. GLAMpoints learns to detect keypoints in a self-supervised manner, exploiting  known spatial correspondence between a specific image and its geometric transformation produced by a controlled homography\footnote{As fundus images depict
small area of retina, it is justified to apply the planar as-
sumption in generating homographies \cite{cattin2006retina,truong2019glampoints}.}. Such full self-supervision has a downside of having many detections on non-vascular areas that are adverse to high-resolution image registration, see \cref{fig:matching}. The non-vascular areas are also unreliable for identity verification. As GLAMPoints is a detector, an external descriptor, \eg rootSIFT \cite{arandjelovic2012three}, is needed. To the best of our knowledge, RIM with jointly trainable keypoint detector and descriptor is non-existing.

\input{fig-exp-matching}

We depart from SuperPoint \cite{detone2018superpoint}, a pivotal work on natural image matching with end-to-end  keypoint detection and description. SuperPoint is a deep network with one encoder followed by two independent decoders. Given a $h \times w$ gray-image input, SuperPoint first uses the encoder to generate  a down-sized feature map of $\frac{h}{8} \times \frac{w}{8} \times 128$. With the feature map as a common input, one decoder produces a full-sized keypoint detection map, while the other decoder produces $256$-dimensional descriptor per pixel on a $\frac{h}{8} \times \frac{w}{8}$ image. Despite its encouraging performance on natural image matching, directly applying SuperPoint for RIM is problematic due to the following issues. First, in order to optimize its descriptor, SuperPoint has to compute hinge losses between all pixels, resulting in a complexity of $O((w\times h)^2)$ for both computation and memory footprint. Such a high complexity significantly limits the input image size, in particular for training, making SuperPoint suboptimal for high-resolution retinal image registration. 
Second, the description loss is computed without taking the detected keypoints into account, making the learned descriptors less  discriminative for disentangling genuine pairs from impostors for identity verification. Lastly, while the loss is computed on the $\frac{h}{8} \times \frac{w}{8} \times 256$ descriptor tensor, the tensor has be upsampled to $h \times w \times 256$ to provide descriptors for keypoints detected at the original size. Such an inherent discrepancy between descriptors used in the training and the inference stages affects the performance, see our ablation study. More recent advances such as R2D2 \cite{r2d2} and NCNet \cite{ncnet} have similar or other issues, as we will  discuss in \cref{sec:related}, motivating us to develop a novel method for RIM.

We propose \emph{SuperRetina}, a  semi-\textbf{Super}vised deep learning method for joint detection and description of keypoints for \textbf{Retina}l image matching.
In contrast to  \cite{detone2018superpoint,truong2019glampoints,r2d2} which limit themselves to fully self-supervised  (without using any manual annotation), we  opt to initialize the training procedure with a relatively small set of (nearly 100) images, sparsely labeled to make the labelling cost well affordable. Such small-scale, incomplete yet precise supervision lets SuperRetina quickly focus on \xredit{specific vascular points such as crossover and bifurcation that are more stable and repeatable}. To overcome the incompleteness of manual labeling, we propose \emph{Progressive Keypoint Expansion} (PKE) to enrich the labeled set at each training epoch. This allows SuperRetina to detect keypoints at previously untouched areas of the vascular tree. Moreover, we modify the network architecture of SuperPoint to directly produce a full-sized descriptor tensor of $h \times w \times 256$, see \cref{fig:network}. Consequently, our description loss is a keypoint-based improved triplet loss, 
which not only leads to highly discriminative descriptors but also has a quadratic complexity \wrt the number of detected keypoints. As this number is much smaller than $h \times w$, SuperRetina allows a  larger input for training. Hence, SuperRetina detects keypoints that are spread over the image plane and at the same time on the vascular tree, making it versatile for multiple RIM tasks.
In sum, our contributions are as follows: \\
$\bullet$ We propose SuperRetina, the first end-to-end method for RIM with jointly trainable keypoint detector and descriptor. \\
$\bullet$  We propose  PKE to address the incompleteness of manual labeling in semi-supervised learning. To enlarge the input size for both training and inference and for highly discriminative descriptors, we re-purpose and adapt a triplet loss as our keypoint-based description loss.\\
$\bullet$ Extensive experiments on two RIM tasks, \ie retinal image registration and retina-based identity verification, show the superior performance of SuperRetina against the previous methods including three dedicated to RIM, \ie PBO \cite{oinonen2010identity}, REMEP \cite{hernandez2020rempe} and GLAMpoints \cite{truong2019glampoints}, and four generic, \ie SuperPoint \cite{detone2018superpoint}, R2D2 \cite{r2d2}, SuperGlue \cite{sarlin2020superglue} and NCNet \cite{ncnet}. Code is available at GitHub\footnote{\href{https://github.com/ruc-aimc-lab/SuperRetina}{\text{https://github.com/ruc-aimc-lab/SuperRetina}}}.

%% file: fig-exp-matching.tex
\begin{figure}[tb!]
  \centering
  \includegraphics[width=1\linewidth]{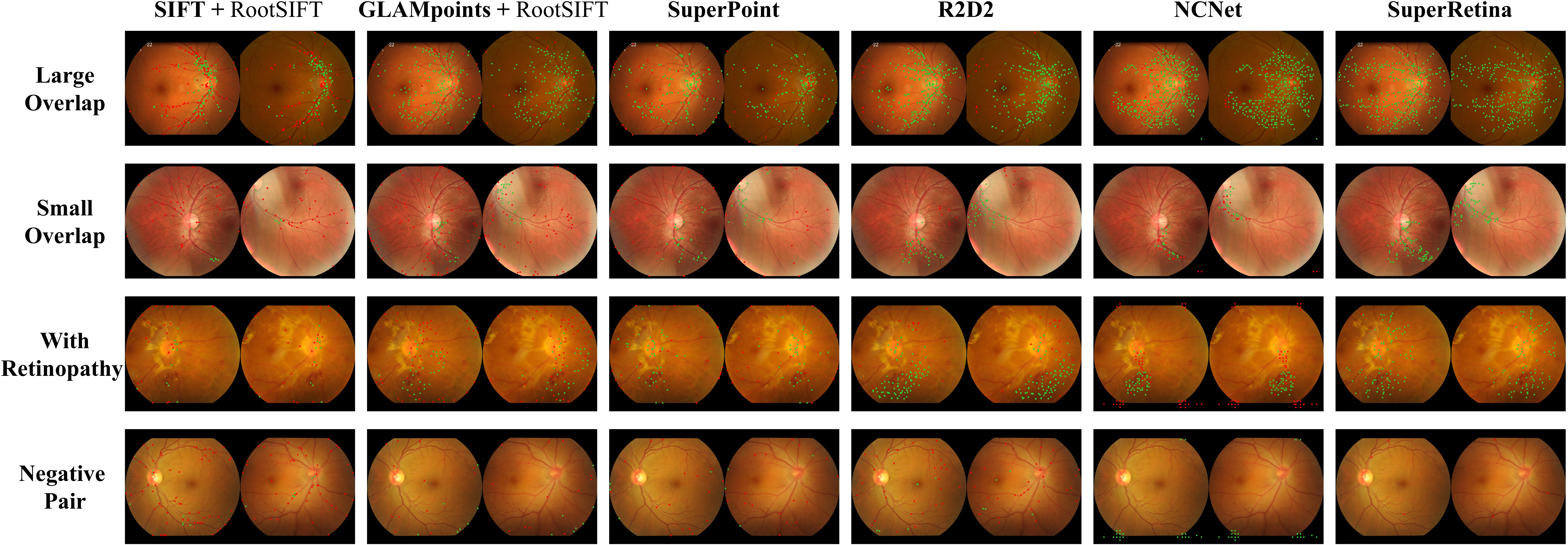}
  \caption{\textbf{Retinal image matching by different methods}. Keypoints corresponding to geometrically valid/invalid matches are shown in green/red dots. The first three rows are positive pairs, \ie retinal images taken from the eye. More green dots and fewer red dots on the positive pairs indicate better matching. For the negative pair, fewer green is better. Best viewed on screen.}
\label{fig:matching}
\end{figure}

%% file: related.tex
\textbf{Progress on Retinal Image Matching}.
Previous works on RIM are tailored to a specific task, let it be single-modal \cite{truong2019glampoints,hernandez2020rempe} or multi-modal \cite{lospa-cvpr15,deepspa-iccv19,wang2020segmentation} image registration, or identity verification \cite{oinonen2010identity,lajevardi2013retina,aleem2018fast}.
For retinal image registration, LoSPA \cite{lospa-cvpr15} and DeepSPA as its deep learning variant \cite{deepspa-iccv19} focus on describing image patches by step pattern analysis (SPA), with keypoints found by detecting intersection points. Designed for feature matching between multi-modal retinal images of the same eye,  the SPA descriptor lacks discrimination in revealing eye identity. GLAMpoints \cite{truong2019glampoints} is trained in a labeling-free manner by exploiting spatial correspondences between a given image and its geometric transformations.
However, such full self-supervision tends to detect many keypoints on non-vascular areas.  
REMPE \cite{hernandez2020rempe} first finds many candidate points by vessel bifurcation detection and the SIFT detector \cite{lowe2004distinctive}, and then performs point pattern matching (PPM) based on eye modelling and camera pose estimation to identify geometrically valid matches. The PPM algorithm involves expensive online optimization, requiring over three minutes to complete a registration, and thus putting its practical use into question.

For identity verification, existing works focus on detecting a few landmarks on the vascular tree, mainly crossover and bifurcation points known to be unique and stable across persons and ages \cite{oinonen2010identity,lajevardi2013retina,aleem2018fast}. With the detected landmarks as input, PPM is then performed. 
PBO \cite{oinonen2010identity} improves PPM by considering principal bifurcation orientations. BGM \cite{lajevardi2013retina} formulates the retinal vasculature as a spatial graph and consequently implements PPM by graph matching. 
Aleem \etal \cite{aleem2018fast}  enhance point patterns of a given image based on spatial relationships between the landmarks, and then vectorize the patterns to a matching template. The number of keypoints required for identity verification is much less than that for image registration.
Probably due to this reason, we see no attempt to re-purpose an identity verification method for image registration. 
In short, while there are few separated efforts on trainable detector (GLAMpoints) and descriptor (DeepSPA) for RIM, a joint effort remains missing. 

\textbf{Progress on Natural Image Matching}. In contrast to RIM, a number of end-to-end methods exist for natural image matching, including SuperPoint \cite{detone2018superpoint}, R2D2 \cite{r2d2}, SuperGlue \cite{sarlin2020superglue}, NCNet \cite{ncnet}, LoFTR \cite{sun2021loftr}, COTR \cite{jiang2021cotr}, PDC-Net \cite{truong2021learning}, \etc. As the newly developed methods focus on natural scenes where detecting repeatable keypoints is difficult due to the lack of repetitive texture patterns, we notice a new trend of keypoint-free image matching.  R2D2 softens the notion of keypoint detection by producing two probalistic maps to measure the reliability and the repeatability per pixel. In NCNet, all pairwise feature matches are computed, resulting in a quadratic complexity \wrt the number of pixels. As a consequence, the feature map used for matching has to be substantially downsized to make the computation affordable. LoFTR improves over SuperGlue with transformers to exploit self-/inter- correlations among the dense-positioned local features. These dense features are powerful for finding correspondences in low-texture areas, desirable for scene image matching. However, this will produce many unwanted matches in non-vasuclar areas when matching retinal images.

%% file: method.tex
SuperRetina is a deep neural network that takes as input a (gray-scale) $h\times w$ retinal image $I$, detects and describes keypoints in the given image with high repeatability and reliability in a single forward pass. We describe the network architecture in  \cref{ssec:network}, followed by the proposed  training algorithms in  \cref{ssec:training}. The use of SuperRetina for RIM is given in \cref{ssec:tasks}.

\subsection{Network Architecture} \label{ssec:network}

We adapt the SuperPoint network. Conceptually, our network consists of an encoder to extract down-sized feature maps $F$ from the given image $I$. The feature map is then fed in parallel into two decoders, one for keypoint detection and the other for keypoint description, which we term Det-Decoder and Des-Decoder, respectively. The Det-Decoder generates a full-sized probabilistic map $P$, where $P_{i,j}$ indicates the probability of a specific pixel being a keypoint, $i=1,\ldots,h$ and $j=1,\ldots,w$. The Des-Decoder produces a $h \times w \times d$ tensor $D$, where $D_{i,j}$ denotes a $d$-dimensional descriptor. Note that in the inference stage, Non-Maximum Suppression (NMS) is applied on $P$ to obtain a binary mask $\widehat{P}$ as the final detection result.  We formalize the above process as follows:
\begin{equation} \label{eq:general}
\left\{ \begin{array}{ll}
 F & \leftarrow \mbox{Encoder}(I), \\
 P & \leftarrow \mbox{Det-Decoder}(F), \\
 D & \leftarrow \mbox{Des-Decoder}(F), \\
 \widehat{P} & \leftarrow \mbox{NMS}(P). \\
\end{array}
\right.
\end{equation}
As illustrated in  \cref{fig:network}, we modify both Det-Decoder and Des-Decoder for RIM.

\textbf{U-Net as Det-Decoder}. 
Effectively capturing low-level patterns such as crossover and bifurcation on the vascular tree is crucial for detecting retinal keypoints in a reliable and repeatable manner. We therefore opt to use U-Net \cite{ronneberger2015u}, originally developed for biomedical image segmentation with its novel design of re-using varied levels of features from the encoder in the decoder by skip connections. In order to support high-resolution input, our encoder is relatively shallow, with a conv layer to generate low-level full-sized feature maps, followed by three conv blocks, each consisting of two conv layers, $2\times2$ max pooling and ReLU. Consequently, the high-level feature maps $F$ have a size of $\frac{h}{8} \times \frac{w}{8} \times 128$.  In order to recover full-sized feature maps, our Det-Decoder uses three conv blocks, each having two conv layers, followed by bilinear upsampling\footnote{We use bilinear upsampling, as transposed convolutions originally used by U-Net are  computationally more expensive, and introduce unwanted checkerboard artifact\cite{M2U-Net}.}, ReLU and concatenation to merge the corresponding feature maps from the encoder. Lastly, a conv. block consisting of three conv. layers and one sigmoid activation is applied on the full-sized feature maps to produce the  detection map $P$.

\textbf{Full-sized Des-Decoder}.
Different from SuperPoint which computes its description loss on a down-sized tensor of $\frac{h}{8} \times \frac{w}{8} \times d$, we target optimizing the descriptors on the full size of $h \times w$, where each pixel is associated with a $d$-dimensional descriptor. Naturally, such dense results are obtained by interpolation, meaning gradient correlation between each keypoint and its neighborhood during backpropagation. Enlarging the neighborhood enhances the correlation, and is thus helpful for training with a larger receptive field \cite{chen2017deeplab}. In that regard, our Des-Decoder first downsizes $F$ to more compact feature maps of $\frac{h}{16}\times \frac{w}{16}\times d$, and then uses an upsampling block (using transposed conv) to generate the full-sized descriptor tensor $D$ of $h \times w \times d$. All the descriptors are $l_2$-normalized.

Our network adaption may seem to be conceptually trivial. Note that producing a full-sized descriptor tensor is computationally prohibitive for a pixel-based description loss as used in SuperPoint and NCNet. A keypoint-based description loss is needed. Nonetheless, keypoint-based training is nontrivial, as inadequate annotations will make the network quickly converge to a local, suboptimal solution. However, having many training images adequately labeled is known to be expensive. To tackle the practical challenge, we develop a semi-supervised training algorithm that works with a small amount of incompletely labeled images.

\input{fig-training-detector}

\subsection{Training Algorithm}  \label{ssec:training}

\textbf{Semi-Supervised Training of Det-Decoder}.
We formulate keypoint detection as a pixel-level binary classification task \cite{detone2018superpoint, truong2019glampoints}. Due to the sparseness and incompleteness of manually labeled keypoints,  training Det-Decoder using a common binary cross-entropy (CE) loss is difficult. To attack the sparseness (and the resultant class imbalance) issue, we leverage two tactics. The first tactic, borrowed from Pose Estimation \cite{wei2016convolutional}, is to convert the binary labels $Y$ to soft labels $\tilde{Y}$ by 2D Gaussian blur, where each keypoint is a peak surrounded by neighbors with their values decaying exponentially. The second tactic is to use the Dice loss \cite{milletari2016v}, found to be  more effective than the weighted CE loss and the Focal loss to handle extreme class imbalance \cite{icpr20-lesion-net}. The Dice-based classification loss $\ell_{clf}$ per image is computed as 
\begin{equation} \label{eq:clfloss}
\ell_{clf}(I; Y)=1 - \frac{2 \cdot \sum_{i,j} {(P  \circ \tilde{Y})_{i,j}}}{\sum_{i,j}{ (P \circ P)_{i,j}}+\sum_{i,j}{(\tilde{Y} \circ \tilde{Y})_{i,j}}},
\end{equation}
where $\circ$ denotes element-wise multiplication.

To attack the incompleteness issue, we propose \textbf{Progressive Keypoint Expansion} (PKE). The basic idea is to progressively expand the labeled keypoint set $Y$ by adding novel and reliable keypoints found by  Det-Detector, which itself is continuously improving after each epoch. To distinguish from such a dynamic $Y$, for each training image we now use $Y_0$ to indicate its initial keypoints, and $S_t$ to denote keypoints detected at the $t$-th epoch, $t=1, 2, \ldots$. We obtain the expanded keypoint set $Y_t$ as $Y_0 \cup S_t$, which \xredit{is used for training at the $t$-th} epoch. 

As $S_t$ is auto-constructed, improper keypoints are inevitable, in particular at the early stage when the Det-Decoder is relatively weak. Given that a good detector shall detect the same keypoint under different viewpoints and scales, GLAMpoints performs a geometric matching to identify keypoints that can be repeatedly detected from a given image and its projective transformations. We improve over GLAMpoints by adding a content-based matching, making it a \emph{double}-matching strategy. As \cref{fig:double-matching} shows, suppose a keypoint detected in a non-vascular area in $I$ (orange circle) has a geometrically matched keypoint (orange square) in $I'=\mathcal{H}(I)$, with $\mathcal{H}$ as a specific homography. Non-vascular areas lack specificity in visual appearance, meaning descriptors extracted such areas are relatively close. Hence, even if the square is the best match to the circle in the descriptor space, it is not sufficiently different from the second-best match to pass Lowe's ratio test \cite{lowe2004distinctive}. Double matching is thus crucial. 

\input{fig-pke-flow}

As illustrated in \cref{fig:pke-flow}, the PKE module works as follows: \\
1) Construct $I^\prime$, a geometric mapping of $I$, using $I^\prime = \mathcal{H}(I)$.\\
2) Feed $I^\prime$ to SuperRetina to obtain its probabilistic detection map $P^\prime$. The inverse projection of the map \wrt $I$ is obtained as  $P^\prime_*=\mathcal{H}^{-1}(P^{\prime})$.  \\
3) Geometric matching: For each point $(i,j)$ in $\widehat{P}$, add it to $S_t$ if ${(P^\prime_*)}_{i,j}>0.5$. \\
4) Content-based matching: For each point $(i,j)$ in $S_t$, we obtain its descriptor by directly sampling the output of the Des-Decoder, resulting in a descriptor set $D_t$. Similarly, we extract $D^\prime_t$ from $I^\prime$ based on $\mathcal{H}(S_t)$. Each descriptor in $D_t$ is used as a query to perform the nearest neighbor search on $D^\prime_t$. A point $(i,j)$ will be preserved in $S_t$, only if its spatial correspondence $(i^\prime,j^\prime)$ passes the ratio test.

The above procedure allows us to progressively find new and reliable keypoints, see \cref{fig:pke-demo}. Moreover, in order to improve the holistic consistency between the detection maps of $I$ and its geometric transformation $I^\prime$, we additionally compute the Dice loss between $P$ and $P^\prime_*$, termed as $\ell_{geo}(I,\mathcal{H})$.  Our detection loss $\ell_{det}$ conditioned on  $Y_t$ and $\mathcal{H}$ is computed as 
\begin{equation} \label{eq:det-loss}
\ell_{det}(I; Y_t, \mathcal{H}) = \ell_{clf}(I; Y_t) + \ell_{geo}(I,\mathcal{H}).    
\end{equation}

\textbf{Self-Supervised Training of Des-Decoder}. 
Ideally, the output of the Des-Decoder shall be invariant to homography. That is, for each keypoint $(i,j)$ detected in $I$, its descriptor shall be identical to the descriptor extracted at the corresponding location $(i^\prime, j^\prime)$ in $I^\prime$. To avoid a trivial solution of yielding a constant descriptor, we choose to optimize a triplet loss \cite{schroff2015facenet} such that the distance between paired keypoints shall be smaller than the distance between unpaired keypoints. Recall that  keypoints are automatically provided by the Det-Decoder, our Des-Decoder is trained in a fully self-supervised manner. Such a property lets the Des-Decoder learn from unlabeled data with ease.

Feeding $I$ and $I^\prime$ separately into SuperRetina allows us to access their full-sized descriptor tensors $D$ and $D^\prime$. For each element $(i,j)$ in the non-maximum suppressed keypoint set $\widehat{P}$, let $D_{i,j}$ be its descriptor. As $(i,j)$ and $(i^\prime, j^\prime)$ shall be paired, the distance of their descriptors, denoted as $\phi_{i,j}$, has to be reduced. With  $(i^\prime, j^\prime)$ excluded, we use $\phi^{rand}_{i,j}$ to indicate the descriptor distance between $(i,j)$ and a point chosen randomly from $\mathcal{H}(\widehat{P})$. Let $\phi^{hard}_{i,j}$ be  the minimal distance. 
We argue that using $\phi^{rand}_{i,j}$ or $\phi^{hard}_{i,j}$  alone as the negative term in the triplet loss is problematic. As the requirement of $\phi_{i,j}<\phi^{rand}_{i,j}$ is relatively easy to fulfill, using $\phi^{rand}_{i,j}$ alone is inadequate to obtain descriptors of good discrimination. Meanwhile, as the network at its early training stage lacks ability to produce good descriptors, using $\phi^{hard}_{i,j}$ exclusively will make the network hard to train. To resolve the issue, we propose a simple  trick by using the mean of  $\phi^{rand}_{i,j}$ and $\phi^{hard}_{i,j}$ as the negative term. Our description loss $\ell_{des}$ is thus defined as
\begin{equation} \label{eq:des-loss}
\ell_{des}(I; \mathcal{H})= \sum_{(i,j) \in \widehat{P}} \max (0, m+\phi_{i,j} -\frac{1}{2}(\phi^{rand}_{i,j}+\phi^{hard}_{i,j})),
\end{equation}
where $m>0$ is a hyper-parameter controlling the margin. Note that $\ell_{des}$ has a quadratic time complexity \wrt the size of $\widehat{P}$, which is much smaller than $h \times w$. Hence, our description loss is much more efficient than its counterpart in SuperPoint, which is quadratic  \wrt   $h \times w$. As such, given the same amount of GPU resources, SuperRetina can be trained on higher-resolution images.

While we describe the training algorithms of Det-Decoder and Des-Decoder separately, they are jointly trained by minimizing the following combined loss:
\begin{equation}
    \ell(I; Y_t, \mathcal{H}) = \ell_{det}(I; Y_t,  \mathcal{H}) + \ell_{des}(I; \mathcal{H}),
\end{equation}
where the homography $\mathcal{H}$ varies per mini-batch.

\subsection{Keypoint-based Retinal Image Matching} \label{ssec:tasks}

Once trained, the use of SuperRetina for RIM is simple. Given a query image $I_q$ and a reference image $I_r$, we feed them separately into SuperRetina to obtain their keypoint probabilistic maps $P_q$ and $P_r$ and associated descriptor tensors $D_q$ and $D_r$. NMS is performed on $P_q$ and $P_r$ to obtain keypoints as $Kp_q$ and $Kp_r$. Recall that $D_q$ and $D_r$ are full-sized, so the corresponding descriptors $desc_q$ and $desc_r$ are fetched directly from the two tensors. Initial matches between $Kp_q$ and $Kp_r$ are obtained by an OpenCV brute-force matcher. The homography matrix $\mathcal{H}$ are then computed using the matched pairs to register $q$ \wrt $r$. As for identity verification, $\mathcal{H}$ is reused to remove outliers. The two images are accepted as \emph{genuine}, \ie from the same eye, if the number of matched points exceeds a predetermined threshold, and \emph{impostor} otherwise. The above process can be written in just a few lines of Python-style code, see the supplement.

%% file: fig-training-detector.tex
\newlength{\twosubht}
\newsavebox{\twosubbox}

\begin{figure}[htb!]
\sbox\twosubbox{%
  \resizebox{\dimexpr\textwidth-1em}{!}{%
    \includegraphics[height=3cm]{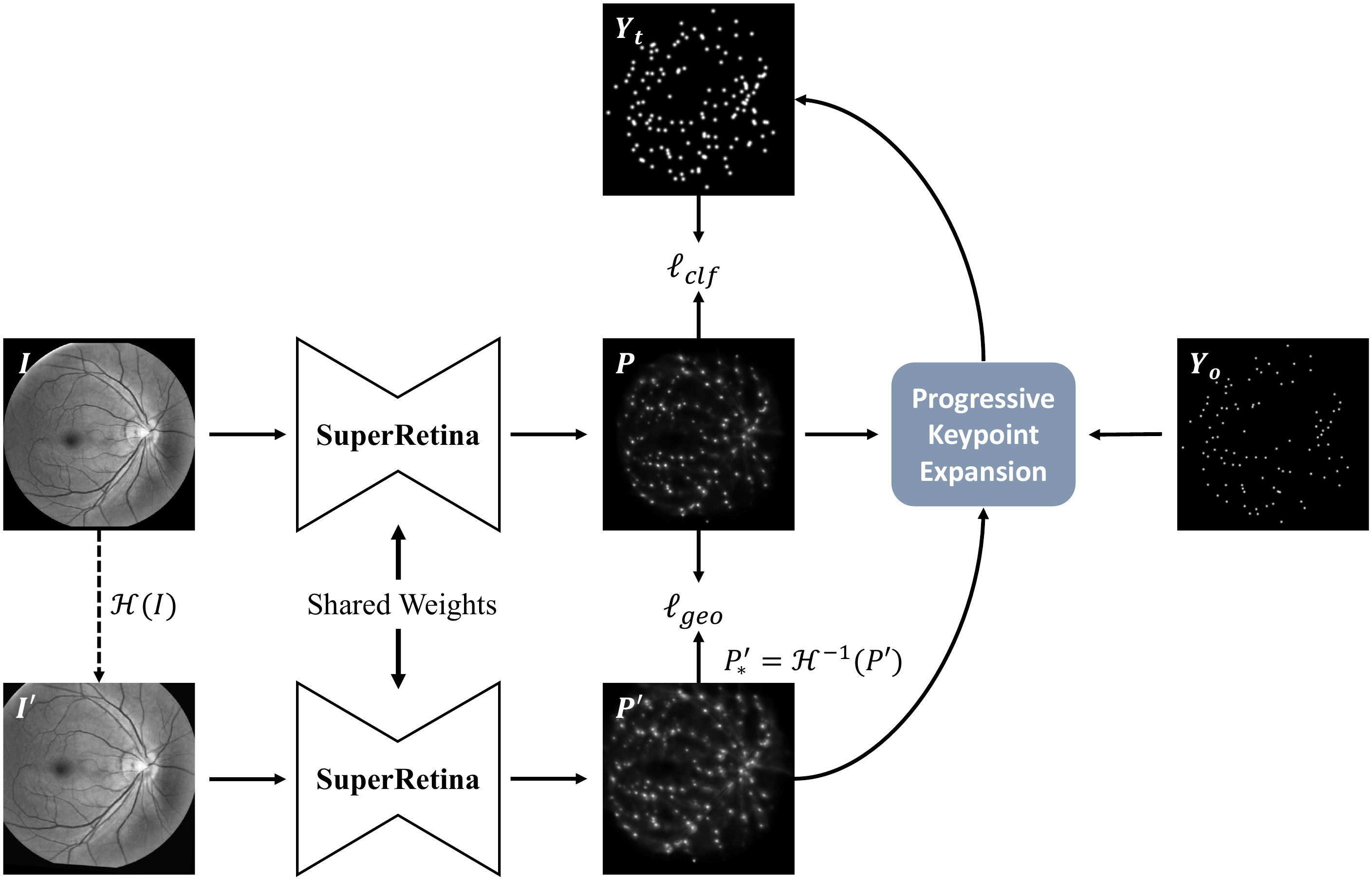}
    \includegraphics[height=3cm]{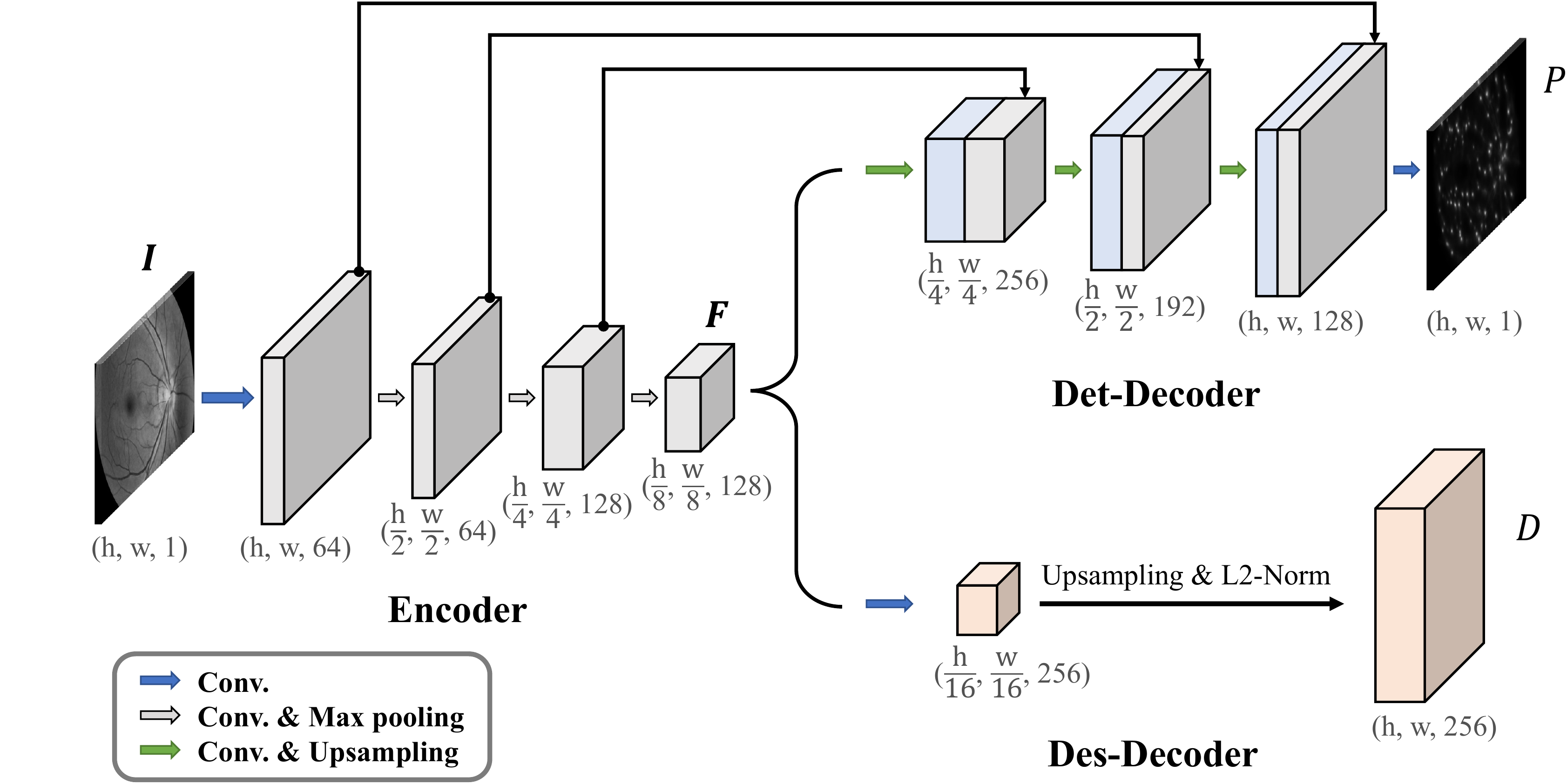}
  }%
}
\setlength{\twosubht}{\ht\twosubbox}


\centering

\subcaptionbox{\label{fig:pke} Training with PKE}{%
  \includegraphics[height=\twosubht]{figures/fig_overall_a.pdf}%
} \quad
\subcaptionbox{\label{fig:network} Network architecture}{%
  \includegraphics[height=\twosubht]{figures/fig_overall_b.pdf}%
}

\sbox\twosubbox{%
  \resizebox{\dimexpr\textwidth-1em}{!}{%
    \includegraphics[height=3cm]{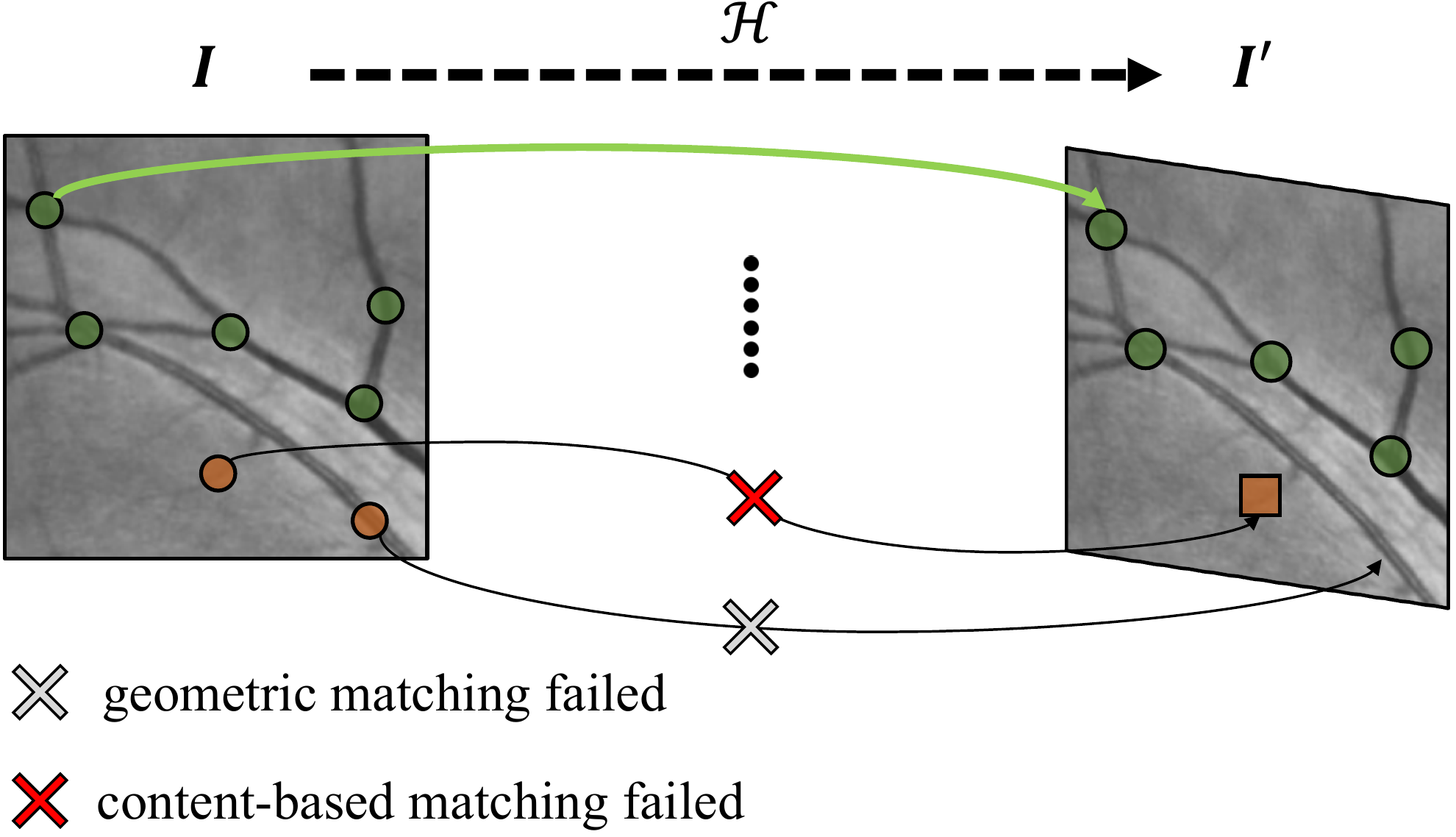}
    \includegraphics[height=3cm]{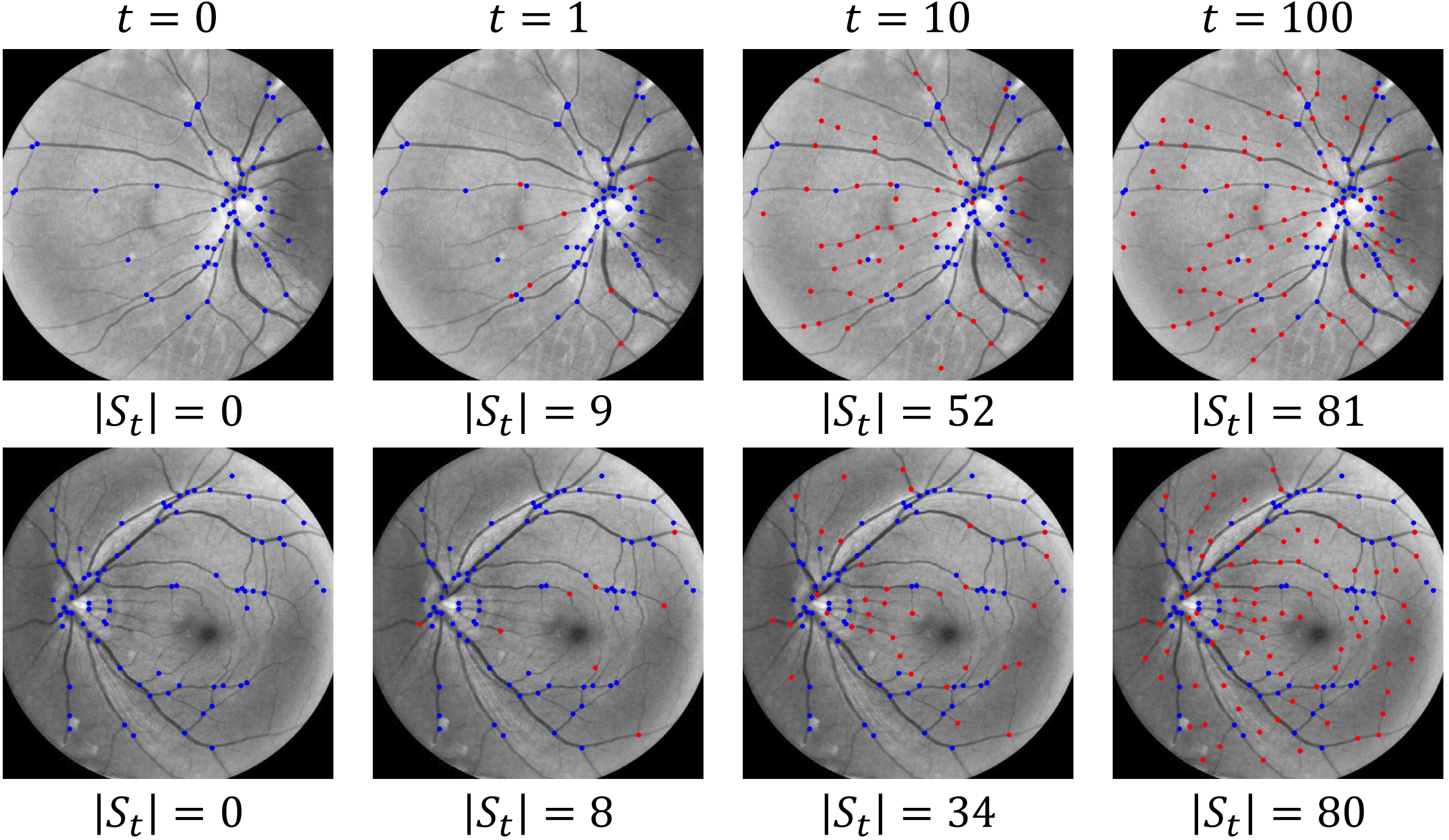}
  }%
}
\setlength{\twosubht}{\ht\twosubbox}

\centering

\subcaptionbox{\label{fig:double-matching} Double-matching strategy}{%
  \includegraphics[height=\twosubht]{figures/fig_overall_c.pdf}%
}
\quad
\subcaptionbox{\label{fig:pke-demo} Newly added keypoints $S_t$ (red dots)}{%
  \includegraphics[height=\twosubht]{figures/fig_overall_d.pdf}%
}
\caption{\textbf{Proposed SuperRetina}. Green/orange markers in (c) indicate genuine/fake keypoints. Blue/red dots in (d) indicate the initial keypoints (auto-detected by PBO \cite{oinonen2010identity}) / iteratively detected keypoints for training.}

\label{fig:training-detector}

\end{figure}

%% file: fig-pke-flow.tex
\begin{figure}[htb!]
  \centering
  \includegraphics[width=0.65\linewidth]{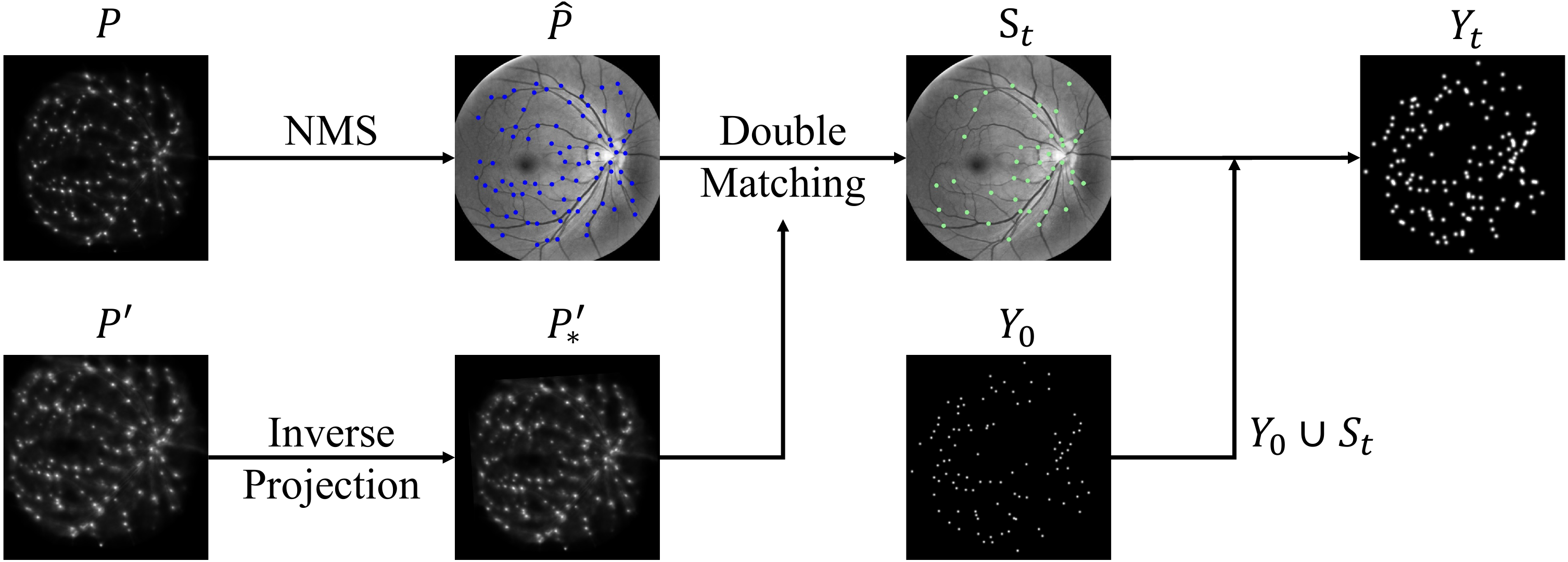}
   \caption{\textbf{Key dataflow within the PKE module}.}
   \label{fig:pke-flow}
\end{figure}

%% file: experiment.tex
To evaluate  SuperRetina in a real scenario, we train it on fixed data. The model is then applied directly (w/o re-training) for different RIM tasks on multiple testsets independent of the training data (\cref{tab:data}).

\subsection{Common Setup} \label{ssec:exp-setp}

\input{exp-setup}

\subsection{Task 1. Retinal Image Registration} \label{ssec:exp-img-reg}
\input{exp-reg}

\subsection{Task 2. Retina-based Identity Verification} \label{ssec:exp-id}

\input{exp-id}

%% file: exp-setup.tex
\textbf{Training data}. We built a small labeled set as follows. We invited 10 members (staffs and students) from our lab. With ages ranging from 22 to 42, the subjects are with normal retinal condition. Multiple color fundus images of the posterior pole (FoV of $45^{\circ}$) were taken per eye, using a SYSEYE Reticam 3100 fundus camera. We collected 97 images in total. The number of keypoints manually labeled\footnote{Keypoint labeling requires little medical knowledge. The first author performed the labeling task in 4 working hours, which we believe was affordable.} per image is between 46 and 147 with a mean value of $93.3$. We term the labeled dataset \emph{Lab}. In addition, to support training of our Des-Decoder, we collected an auxiliary dataset of 844 retinal images from 120 subjects having varied retinal diseases. Recall that Des-Decoder is trained in a fully self-supervised manner, so the auxiliary dataset requires no extra annotation.

\textbf{Implementation}. We implement SuperRetina using PyTorch.
Subject to our GPU resource (an NVIDIA GeForce RTX 2080 Ti), we choose a training input size of~$768 \times 768$. The network is trained end-to-end by SGD with mini-batch size of 1. The optimizer is Adam \cite{kingma2014adam}, with $\beta=(0.9, 0.999)$ and an initial learning rate of $0.001$. Standard data augmentation methods are used: gaussian blur, changes of contrast, and illumination. The number of maximum training epochs is 150. The descriptor length $d$ is 256. For inference, the NMS size is 10$\times$10 pixels. For homography fitting, we use \texttt{cv2.findHomography} with LMEDS.

\input{tables/table_dataset}



%% file: tables/table_dataset.tex
\begin{table}[tbh!]
\centering
\ra{1}
\caption{\textbf{Our experimental data}. Large cross-dataset divergence \wrt subjects, retinal conditions, imaging FoV \etc allows us to evaluate the effectiveness and generalization ability of SuperRetina. All test images are resized to  $768\times768$, except for images from VAIRA which use $512\times512$ due to their smaller FoV.}
\resizebox{0.7\linewidth}{!}{

\begin{tabular}{@{}l|r|r|r|r|r|r@{}}

\toprule
\multirow{2}{*}{\textbf{Dataset}} & \multicolumn{1}{r|}{\multirow{2}{*}{\textbf{Subjects}}} & \multicolumn{1}{r|}{\multirow{2}{*}{\textbf{Eyes}}} & \multicolumn{1}{r|}{\multirow{2}{*}{\textbf{Images}}} & \multicolumn{3}{c}{\textbf{Image pairs}}\\ 
\cmidrule{5-7} 
& & & & \multicolumn{1}{r|}{\textit{Total}} & \multicolumn{1}{r|}{\textit{Genuine}} & \multicolumn{1}{r}{\textit{ Impostor}} \\ 
\midrule
\multicolumn{7}{@{}l}{\textit{Training sets:}}\\
Lab (labeled) & 10  & 20  & 97 & -- & -- & --\\  
Auxiliary (unlabeled) & 120 & 215 & 844 & -- & -- & -- \\ [3pt] 

\multicolumn{7}{@{}l}{\textit{Test set for retinal image registration}}\\ FIRE\cite{hernandez2017fire}  & -- & -- & 129 & 134 & 134 & -- \\  [3pt]

\multicolumn{7}{@{}l}{\textit{Test sets for retina based identity verification}}\\ 
VARIA \cite{ortega2009retinal} & -- & 139 & 233 & 27,028 & 155 & 26,873 \\ 
CLINICAL & 100 & 180 & 691 & 16,203 & 1,473 & 14,730 \\ 
BES \cite{BES1, BES2} & 2,066 & 4,132 & 24,880 & 99,846 & 49,923 & 49,923 \\

\bottomrule
\end{tabular}
}

\label{tab:data}

\end{table}

%% file: exp-reg.tex
\textbf{Test set}.
We adopt FIRE  \cite{hernandez2017fire}, a benchmark set consisting of 129 images of size $2,912 \times 2,912$ acquired with a Nidek AFC-210 fundus camera (FOV of $45^{\circ}$) and 134 registered image pairs. The pairs have been divided into three groups according to their registration difficulty: \emph{Easy} (71 pairs with high overlap and no anatomical change), \emph{Moderate} (14 pairs with high overlap and large anatomical changes), and \emph{Hard} (49 pairs with small overlap and no anatomical changes).

\textbf{Performance metrics}. 
Following \cite{truong2019glampoints}, we report three sorts of rates, \ie failed, inaccurate and acceptable. Given a query image $I_q$ and its reference $I_r$, a registration is considered failed if the number of matches is less than $4$,  the minimum required to estimate a homography $\mathcal{H}$. Otherwise, for each query point $p_q$ in $I_q$, we compute the $l_2$ distance between $\mathcal{H}(p_q)$ and its reference $p_r$ in $I_r$. Per query image, the median distance is defined as the median error (MEE), with the maximum distance as the maximum error (MAE). A registration is considered acceptable if $\mbox{MEE}<20$ and $\mbox{MAE}<50$, and inaccurate otherwise. Besides, we report Area Under Curve (AUC) proposed by \cite{hernandez2017fire}, which estimates the expectation of the acceptance rates \wrt the decision threshold, and thus reflects the overall performance of a specific method. Following \cite{hernandez2017fire}, we compute AUC per category, \ie Easy, Mod and Hard, and take their mean (mAUC) as an overall measure. Higher acceptance rate / AUC and lower inaccurate / failed rates are better. All the metrics are computed on the original size of $2912\times2912$.

\textbf{Baselines}. For a reproducible comparison, we choose competitor methods that have either source code or pre-trained models released by paper authors. Accordingly, we have \xredit{eight} baselines as follows: \\
$\bullet$ SIFT detector \cite{lowe2004distinctive} plus  RootSIFT descriptor \cite{arandjelovic2012three}, using OpenCV APIs. \\
$\bullet$ PBO \cite{oinonen2010identity}, a traditional keypoint extraction and matching method with author-provided Matlab implementation. \\ 
$\bullet$ REMPE \cite{hernandez2020rempe}, performing retinal image registration through eye modelling and pose estimation\footnote{\href{https://projects.ics.forth.gr/cvrl/rempe/}{\text{https://projects.ics.forth.gr/cvrl/rempe/}}}. \\
$\bullet$ SuperPoint\footnote{\href{https://github.com/rpautrat/SuperPoint}{\text{https://github.com/rpautrat/SuperPoint}}\label{superpoint}} \cite{detone2018superpoint} trained on MS-COCO \cite{lin2014microsoft}. \\
$\bullet$ GLAMpoints\footnote{\href{https://github.com/PruneTruong/GLAMpoints\_pytorch}{https://github.com/PruneTruong/GLAMpoints\_pytorch}\label{glampoints}} \cite{truong2019glampoints} (+ RootSIFT  descriptor) trained on private fundus images. \\
$\bullet$ R2D2\footnote{\href{https://github.com/naver/r2d2}{https://github.com/naver/r2d2}\label{R2D2}} \cite{r2d2}, trained on the Aachen dataset \cite{sattler2018benchmarking}. \\
$\bullet$  SuperGlue\footnote{\href{https://github.com/magicleap/SuperGluePretrainedNetwork}{https://github.com/magicleap/SuperGluePretrainedNetwork\label{superglue}}} \cite{sarlin2020superglue}, trained on ScanNet \cite{dai2017scannet}. \\
$\bullet$ NCNet\footnote{\href{https://github.com/ignacio-rocco/ncnet}{\text{https://github.com/ignacio-rocco/ncnet}}\label{ncnet}} \cite{ncnet}, pretrained on the Indoor Venues Dataset \cite{rocco2018neighbourhood}. 

Due to the natural domain gap between retinal images and natural images, the baseline models pretrained on natural images might not be in their optimal condition for RIM. We take this into account by finetuning SuperPoint, GLAMpoints, R2D2 and NCNet on our training data.

\textbf{Comparison with the Existing Methods}. 
As shown in \cref{tab:exp_all}, SuperRetina, with zero failure, an inaccurate rate of 1.49\% and an acceptance rate of 98.51\% is the best. Interestingly, we find that REMPE, which relies on traditional image processing enhanced by geometric modeling of the retina, performs better than the deep learning based alternatives including GLAMpoints, R2D2, SuperPoint, SuperGlue and NCNet. SuperRetina beats this strong baseline.

\input{tables/table_eval_all}

Similar results are observed in terms of AUC scores. The only exception is on the Easy group, where REMPE obtains a higher AUC (0.958 versus 0.940).  Recall that images in this group have large overlap and no anatomic change, so the heavy modeling of the retinal structure in REMPE is advantageous. The benefit of end-to-end learning becomes more evident when dealing with the Moderate and Hard groups. SuperRetina scores a substantially higher AUC-Mod than REMPE (0.783 versus 0.660). Moreover, while REMPE takes 198 seconds to perform one registration, SuperRetina is far more efficient, requiring 1 second, most of which is spent on data IO and preprocessing. As only the query image has to be computed on the fly, while images in the database can be precomputed, the entire image matching process can be much accelerated. In short, the advantage of SuperRetina over REMPE is three-fold: (i) The end-to-end learned detector is more reliable than REMPE's vessel bifurcation detector for handling images with large anatomical changes, (ii) SuperRetina works for both image registration and identity verification, and (iii) SuperRetina is nearly 200x faster.

\textbf{Manual Labeling versus Auto-Labeling for $Y_0$}. 
The last three rows of \cref{tab:exp_all} are SuperRetina with distinct choices of the initial keypoint set $Y_0$.
Pretraining means we tried to first train SuperRetina on the synthetic corner dataset as used by SuperPoint, and then use this pre-trained SuperRetina to produce $Y_0$. The second-last row means using PBO-detected keypoints as $Y_0$. 
Their results show that even with the auto-produced $Y_0$, SuperRetina compares favorably against the current methods. In particular, using PBO-based $Y_0$ obtains mAUC of $0.750$.
The number, although lower than using the manual $Y_0$ (mAUC $0.755$), clearly outperforms the best baseline, \ie REMPE (mAUC $0.720$). At the cost of merely 0.66\% relative loss in performance, SuperRetina can indeed be trained in a manual-annotation free manner.

\textbf{Evaluating the Influence of PKE}. As \cref{tab:eval_self} shows, SuperRetina w/o PKE suffers from a clear performance drop. Without PKE, the average number of keypoints detected by SuperRetina is substantially reduced, from 530 to 109 per image. We also tried PKE without content-based matching, making it effectively the keypoint selection strategy used by GLAMpoints. Its lower performance (row\#3 in \cref{tab:eval_self}) verifies the necessity of the proposed double-matching strategy. The above results justify the effectiveness of PKE for expanding the annotation data for semi-supervised learning.

For the description loss, we simultaneously leverage the hard negative instance and a random negative for computing the negative term in \cref{eq:des-loss}. We tried an alternative strategy of semi-hard negative sampling, where the negative ranked at the middle among all
candidate negatives in a given mini-batch is chosen for computing the negative term. This alternative strategy (row\#4 in \cref{tab:eval_self}) is ineffective. 

In addition, we re-run the same training pipeline, but \textit{w/o} descriptor upsampling, \textit{w/o} 2D Gaussain blur and using the (weighted) CE loss instead of Dice, respectively. Their consistent lower performance supports the necessity of the proposed changes regarding the network and its training strategy.

\input{tables/table_eval_ablation}

%% file: tables/table_eval_all.tex
\begin{table}[!]
    \ra{1.1}
   \centering
 \caption{\textbf{Performance of the state-of-the-art for two RIM tasks, \ie retinal image registration and retina based identity verification}. Methods postfixed with \emph{finetune} have been finetuned on our training data. The proposed SuperRetina compares favorably against the existing methods, even with the initial keypoint set $Y_0$ automatically detected by the PBO method.  }
\resizebox{\linewidth}{!}{

   \begin{threeparttable}[htbp]

         \begin{tabular}{@{}lrrrrrrrrrrr@{}}
         \toprule
         \multirow{2}*{\centering\textbf{Methods}}
         & \multicolumn{7}{c}{\specialcell{\textbf{Image Registration} \\ (FIRE as the test set)}} &  \phantom{abc} &
         
         \multicolumn{3}{c}{\specialcell{\textbf{Identity Verification} \\(EER [$\%$])}}
         \\
         \cmidrule{10-12}
         \cmidrule{2-8}
         
           & Failed [$\%$] &
             Inaccurate [$\%$] &
             Acceptable [$\%$] &
             AUC-Easy &
             AUC-Mod &
             AUC-Hard &
             mAUC &&
             VARIA  & CLINICAL	& BES
         \\

         \midrule
         \emph{Traditional:} \\
         SIFT, IJCV04 \cite{lowe2004distinctive}  & 0 & 20.15 & 79.85 &  0.903 & 0.474 & 0.341 & 0.573  & & 
         0.65 & 3.64 & 4.67  \\
         PBO, ICIP10 \cite{oinonen2010identity}  & 0.75 & 28.36 & 70.89 & 0.844 & 0.691 & 0.122 & 0.552 &&   0.65 & 4.96 & 4.33\\
         REMPE, JBHI20\cite{hernandez2020rempe}  & 0 & 2.99 & 97.01 & \textbf{0.958} & 0.660 & \textbf{0.542} & 0.720  &  & -- & -- & --  \\ [3pt]
         
         \multicolumn{5}{@{}l}{\emph{Deep learning based:}} \\
         SuperPoint, CVPRW18 \cite{detone2018superpoint} & 0 & 5.22 & 94.78 & 0.882 & 0.649 & 0.490  & 0.674 & &   0.01 & 1.06 & 2.00\\
         SuperPoint-\emph{finetune} & 0 & 6.72 & 93.28  & 0.909 & 0.609 & 0.465 & 0.661  &&  0.01 & 2.89 & 3.91\\
          
         GLAMpoints, ICCV19\cite{truong2019glampoints} & 0 & 7.46 & 92.54 & 0.850 & 0.543 & 0.474  & 0.622 &  &  0.02 & 4.32 & 2.95 \\
         GLAMpoints-\emph{finetune} &  0 & 7.46 & 92.54 & 0.825 & 0.517 & 0.490 & 0.611 &  &  0.03 & 6.74 & 4.83 \\
         R2D2, NIPS19 \cite{r2d2} & 0 & 12.69 & 87.31 & 0.900 & 0.517 & 0.386  & 0.601&  &  0.05 &  6.23 & 7.16 \\
         R2D2-\emph{finetune} & 0 & 4.48 & 95.52 & 0.928 & 0.666 & 0.540& 0.711 &  &  0.05 & 1.83  & 7.76 \\
         SuperGlue, CVPR20 \cite{sarlin2020superglue} & 0.75 & 3.73 & 95.52 & 0.885 & 0.689 & 0.488 & 0.687  &&  0 & 2.38 & 2.35\\
         NCNet, TPAMI22 \cite{ncnet} & 0 & 37.31 & 62.69 & 0.588 & 0.386 & 0.077 & 0.350  &&  14.19 & 22.13 & 30.67\\
         NCNet-\emph{finetune} & 0 & 14.18 & 85.82 & 0.817 & 0.609 & 0.410 & 0.612   & &  7.97 & 3.05 & 19.87
         \\[3pt]
         \textbf{SuperRetina} \\
         $Y_0$: Pretraining & 0 & 2.99  & 97.01   & 0.922 & 0.720 &  0.502 &  0.715  && 0 & 1.04  & 1.93 \\
         $Y_0$: PBO& 0 & 3.73  & 96.27  & 0.944 & \textbf{0.789} & 0.516 & 0.750 & & 0 & 1.02  & \textbf{1.10} \\
         $Y_0$: Manual labeling & 0 &  \textbf{1.49} & \textbf{98.51}  & 0.940 & 0.783 & \textbf{0.542} & \textbf{0.755}  & &   0 & \textbf{0.83} & 1.18\\
         \bottomrule
         \end{tabular}
     
   \end{threeparttable}
    }

   \label{tab:exp_all}
\end{table}

%% file: tables/table_eval_ablation.tex
\begin{table}[htb]
   \ra{1}
   \centering
       \caption{\textbf{Ablation study}. Larger mAUC on FIRE and lower EER on VARIA, CLINICAL and BES are better. }
   \setlength{\abovecaptionskip}{1.5pt}
\resizebox{0.75\columnwidth}{!}{

     \begin{tabular}{@{}lrrrr@{}}
     \toprule
     \textbf{Setup}  & {\textbf{FIRE}($\uparrow$)} & {\textbf{VARIA}($\downarrow$)} &{\textbf{CLINICAL}($\downarrow$)} & {\textbf{BES}($\downarrow$)} \\
     \midrule
     
     Full-setup & 0.755 & 0 & 0.83 & 1.18  \\
      \midrule
     \emph{w/o} PKE & 	0.685 &  	0.01  & 	5.14  & 	3.11 \\
    PKE \emph{w/o} content-based mathcing	 & 0.670  & 	0  & 	1.48  & 	1.19 \\
    Semi-hard negative sampling	 & 0.407  & 	2.75  & 	10.18  & 	7.83 \\
    
    w/o upsampling & 0.697 & 0.03 & 3.46 & 4.15  \\
    w/o Gaussian blur & 0.574 & 8.38 & 7.44 & 10.82  \\
    Dice $\rightarrow$ CE & 0.653 & 0.65 & 4.20 & 2.48 \\
    Dice $\rightarrow$ weighted CE & 0.704 & 0.02 & 1.79 & 1.32 \\ 
	 \midrule
    \multicolumn{5}{@{}l}{\textit{Compare with other detectors:}}\\				
    Det: SIFT, Des: SuperRetina	 & 0.585 & 	0  & 	4.40  & 	4.23 \\
    Det: GLAMpoints, Des: SuperRetina & 	0.605 & 	0  & 	2.84  & 	1.51 \\
    Det: SuperPoint, Des: SuperRetina & 	0.673 &	0  & 	1.60  & 	1.68 \\ 
     \midrule
    \multicolumn{5}{@{}l}{\textit{Compare with other descriptors:}}\\	
    			
    Det: SuperRetina, Des: RootSIFT & 	0.705 & 	0 & 	2.81 & 	2.10\\
    Det: SuperRetina, Des: SOSNet & 	0.712 & 	0 & 	0.88 & 	1.78\\
     
     \bottomrule
     \end{tabular}
 
    }
     \label{tab:eval_self}
\end{table}

%% file: exp-id.tex
 \textbf{Test Sets}. We use three test sets: VARIA \cite{ortega2009retinal}, Beijing Eye Study (BES) \cite{BES1,BES2}, and a private set. VARIA has 233 gray-scale retinal images from 139 eyes, acquired with a Topcon NW-100 camera. The images are  optic disc centered, with a small FoV of around $20^{\circ}$. BES, acquired for a population-based study conducted in Beijing between 2001 and 2011, has 24,880 color fundus photos taken from 4,132 eyes at different periods. As images taken at earlier periods were digital scans of printed photos, the image quality of BES varies. Our private set, termed CLINICAL, consists of 691 images from 100 patients, acquired with a Topcon Trc-Nw6 fundus camera at an outpatient clinic of ophthalmology with due ethics approval. CLINICAL exhibits more diverse abnormal conditions such as old macula lesion, retinitis pigmentosa and macular edema.  The joint use of the testsets leads to a systematic evaluation covering retinas in normal (VARIA)/abnormal (CLINICAL) conditions and across ages (BES).

 \textbf{Performance metric}. We report Equal Error Rate (EER). As a common metric for evaluating a biometric system, EER is the value when the system's False Accept Rate and False Reject Rate are equal. Lower is better.

\textbf{Baselines}. 
We re-use the baselines from  \cref{ssec:exp-img-reg} except for REMPE \cite{hernandez2020rempe}, which is inapplicable for identity verification. 

\textbf{Comparison with State-of-the-Art}.
As \cref{tab:exp_all} shows, SuperRetina, with EER of 0$\%$ on VARIA, $0.83\%$ on CLINICAL and $1.18\%$ on BES, compares favorably against the baselines. All the deep learning based methods perform well on VARIA, which has a small FoV with clearly visible vessels. However, their performance decreases noticeably on CLINICAL and BES, especially for GLAMpoints and R2D2, both using self-supervised training. As shown in 
\cref{fig:matching}, GLAMpoints and R2D2 tend to detect keypoints on non-vascular areas. 
By contrast, SuperRetina keypoints are mostly distributed along the vascular tree, thus more suited for identity verification.

\textbf{Ablation Study}.
\cref{tab:eval_self} shows that PKE also matters for identity verification. As for the choice of $Y_0$, using the PBO-produced labels achieves comparable results for two out of the three test sets, \ie VARIA and BES. Note that its higher EER of $1.02\%$ on CLINICAL remains better than the best baseline, \ie SuperPoint with EER of $1.06\%$.  
We compare the SuperRetina detector with three existing detectors, \ie SIFT, SuperPoint and GLAMpoints, all using the SuperRetina descriptor. We also compare the SuperRetina descriptor with two existing descriptors, \ie RootSIFT previously used by GLAMPoints for RIM and SOSNet, a widely used deep descriptor \cite{tian2019sosnet}. \cref{tab:eval_self} shows that our detector and descriptor remain competitive even used separately.

%% file: supplementary/supplementary.tex







\definecolor{commentcolor}{RGB}{85,139,78}
\definecolor{stringcolor}{RGB}{206,145,108}
\definecolor{keywordcolor}{RGB}{0,0,0}
\definecolor{backcolor}{RGB}{220,220,220}
\SetAlCapNameFnt{\small}
\SetAlCapFnt{\small}

\lstset{						
	language=python, 					
	basicstyle=\fontsize{10}{11}\selectfont\ttfamily,
	linewidth=0.9\linewidth,      		
	commentstyle=\color{commentcolor},	
	keywordstyle=\color{keywordcolor},	
	showspaces=false,					
	numberstyle=\tiny\emptyaccsupp,		
	numbersep=5pt,						
	frame=single,						
	framerule=0pt,						
	escapeinside=@@,					
	emptylines=1,						%
	xleftmargin=-2em,					
	tabsize=4,							
	gobble=4							
}

\section*{Appendix}

\appendix




In this supplementary material, we provide more details of our evaluation, which are not included in the paper due to space limit. 

\textbf{Qualitative comparison between detector-based methods}. \cref{fig:det_res} shows that GLAMpoints and R2D2 tend to detect keypoints in non-vascular areas which are not discriminative for retinal image matching. By contrast, the keypoints found by SuperRetina are mostly distributed along the vascular tree, thus more suited for retinal image matching.

\input{supplementary/fig-keypoint}

\textbf{Performance curves of different methods for retinal image registration}.  \cref{fig:auc} shows performance curves of the registration successful rate \wrt the error threshold at varied levels in three distinct modes, \ie \emph{Easy}, \emph{Moderate} and \emph{Hard}.
We plot the curves using the source code  kindly provided by the developers of the FIRE dataset \cite{hernandez2017fire, hernandez2020rempe} via personal correspondence.

\input{supplementary/fig-auc}

\textbf{DET graphs of different methods for retina-based identity verification}. \cref{fig:det} shows the Detection Error Tradeoff (DET) graphs for the identity verification task.

\input{supplementary/fig-eer}


\textbf{Pseudo-code of using a trained SuperRetina model for retinal image matching}. 
The use of SuperRetina for RIM mentioned in Section 3.3 can be written in just a few lines of  Python-style code, see Algorithm \ref{algorithm1}.

\textbf{Other features (lesion / optic disc) for RIM}?  We do not consider lesions and optic disc. While lesions are stable in a specific examination session, they can be unstable across sessions, e.g., reduced after proper treatment or growing due to disease progress. Lesions are not eye-specific, and thus unsuited for identity verification. Moreover, unlike vascular keypoints, labeling lesions requires retinal expertise, making them nontrivial to obtain \cite{icpr20-lesion-net}. The optic disc is  not eye-specific also. Moreover, for retinal images of the posterior pole (FoV of $45^{\circ}$), the optic disc possesses a relatively small percentage of the FoV area (1/36), making it less significant for matching. That said, our method is generic and can in principle be used to detect other features by instantiating $Y_0$ with keypoints related to those features. 

\begin{algorithm}[H]
  \caption{SuperRetina for multi-task RIM}\label{algorithm1}
  \begin{lstlisting}[language=python] 
      # q, r: query and reference images
      # thresh: acceptance threshold

      # Detect and describe keypoints
      P_q, D_q = SuperRetina(q) 
      P_r, D_r = SuperRetina(r)
      # NMS to get keypoints
      Kp_q = NMS(P_q)
      Kp_r = NMS(P_r)
      # Sample descriptions
      desc_q = sample_desc(D_q, Kp_q)
      desc_r = sample_desc(D_r, Kp_r)
      # Keypoint match using Brute-force matcher
      matches = bfMatch(desc_q, desc_r)
      if len(matches) < 4:
         reject and exit  # matching failed
      # Compute H for image registration   
      H = findHomography(Kp_q, Kp_r, matches, cv.LMEDS)
      # Remove outlier matches for identity verification
      matches = remove_outliers(matches, H)
      accept = (len(matches) >= thresh)
  \end{lstlisting}
   
\end{algorithm}

\clearpage
%
%

%% file: supplementary/fig-keypoint.tex
\begin{figure} [h!]
   \begin{center}
   \includegraphics[width=\columnwidth]{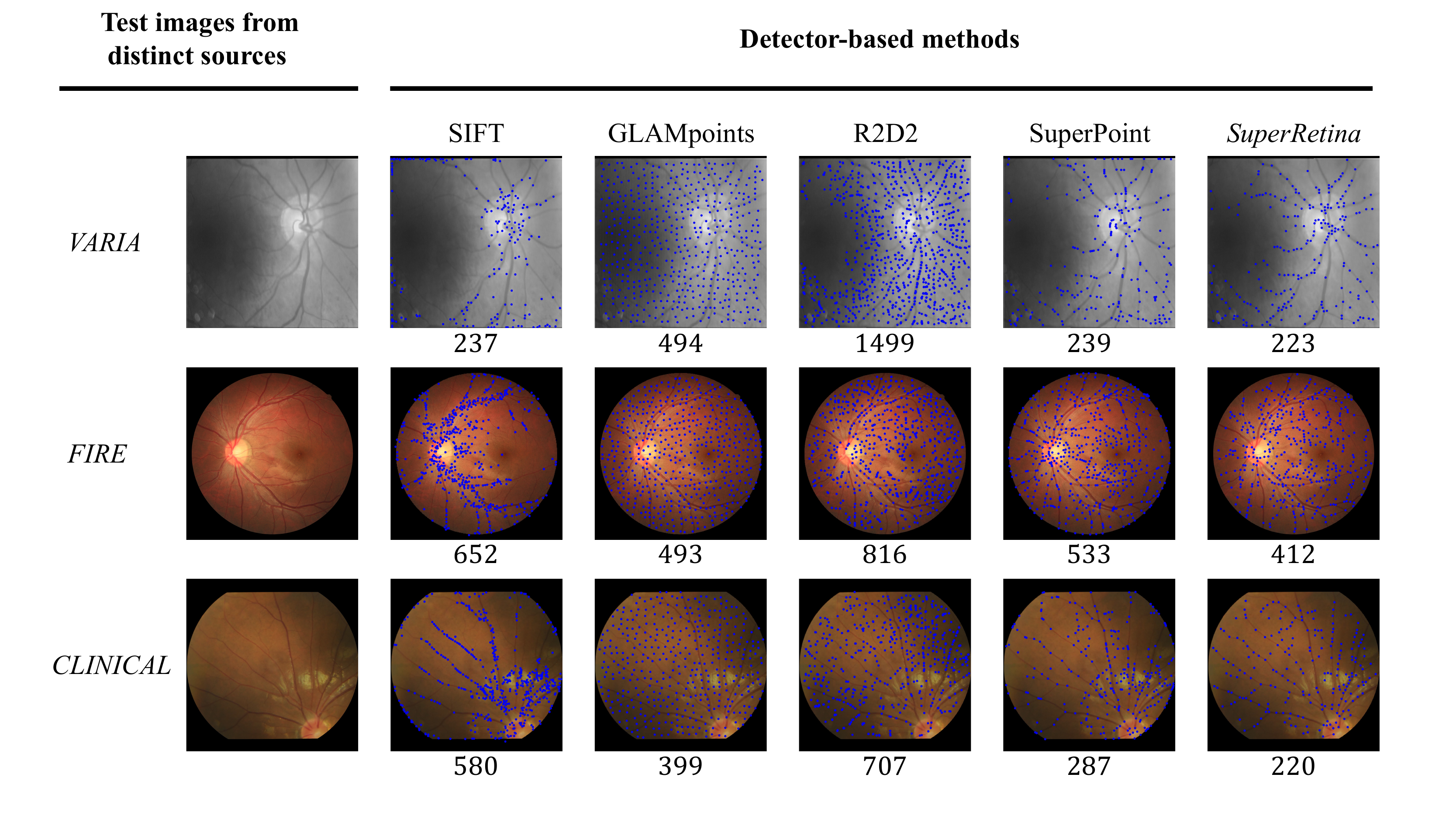}
   \end{center}
      \caption{\textbf{Detected keypoints in retinal images from distinct data sources}, \ie  VARIA \cite{ortega2009retinal}, FIRE \cite{hernandez2017fire} and CLINICAL, a private dataset collected in  clinical scenarios by this paper. Numbers below each image are the amount of keypoints found by a specific method. The proposed SuperRetina detects keypoints that spread over the field-of-view and in the meantime fall on the vascular tree.}
   \label{fig:det_res}
\end{figure}

%% file: supplementary/fig-auc.tex
\begin{figure}[h!]
    \centering
        \begin{subfigure}{0.32\columnwidth}
         \includegraphics[width=\columnwidth]{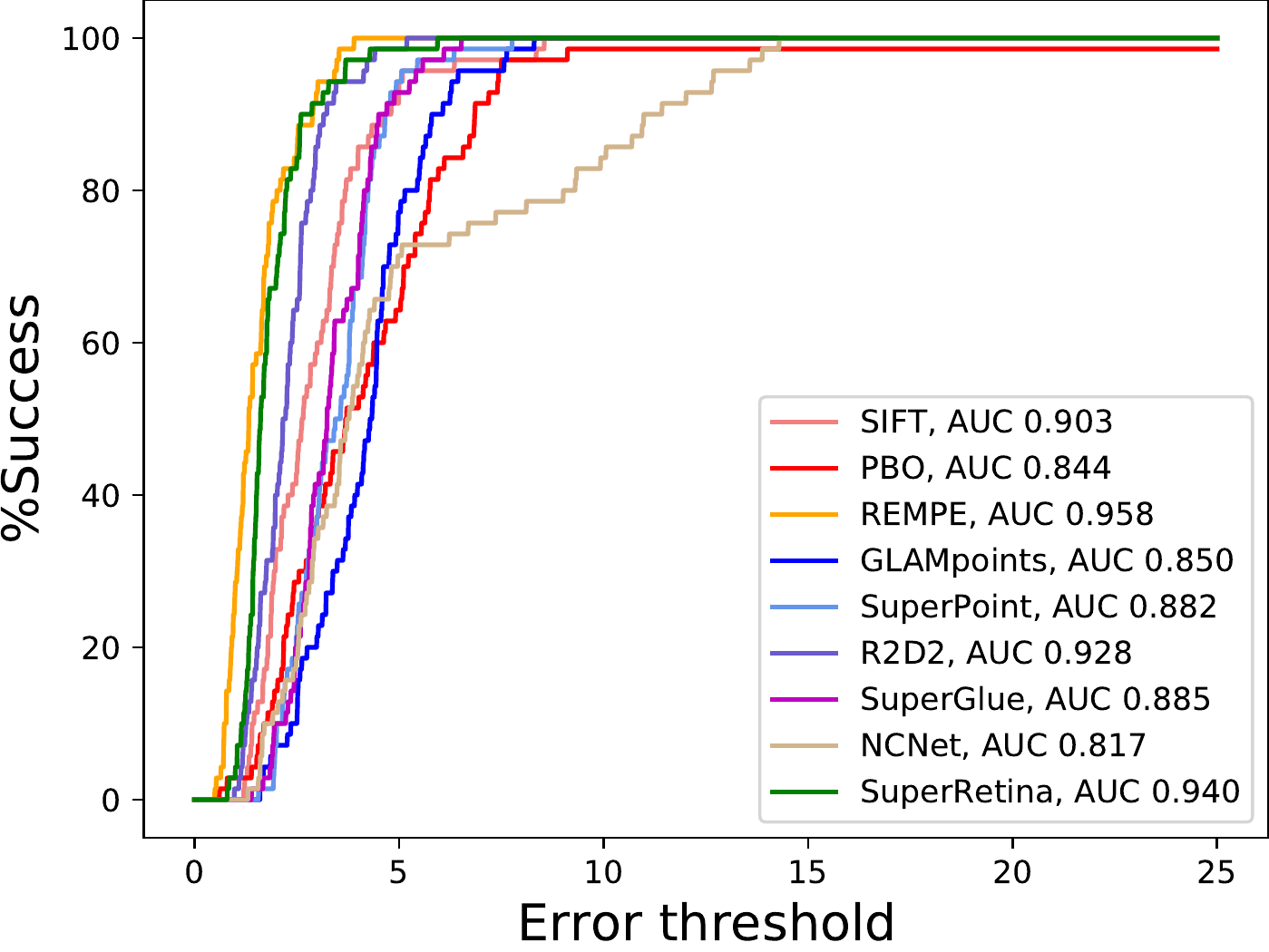}
         \caption{Easy mode}
         \label{fig:auc-sota-Easy}
         \end{subfigure}
         \begin{subfigure}{.32\columnwidth}
         \includegraphics[width=\columnwidth]{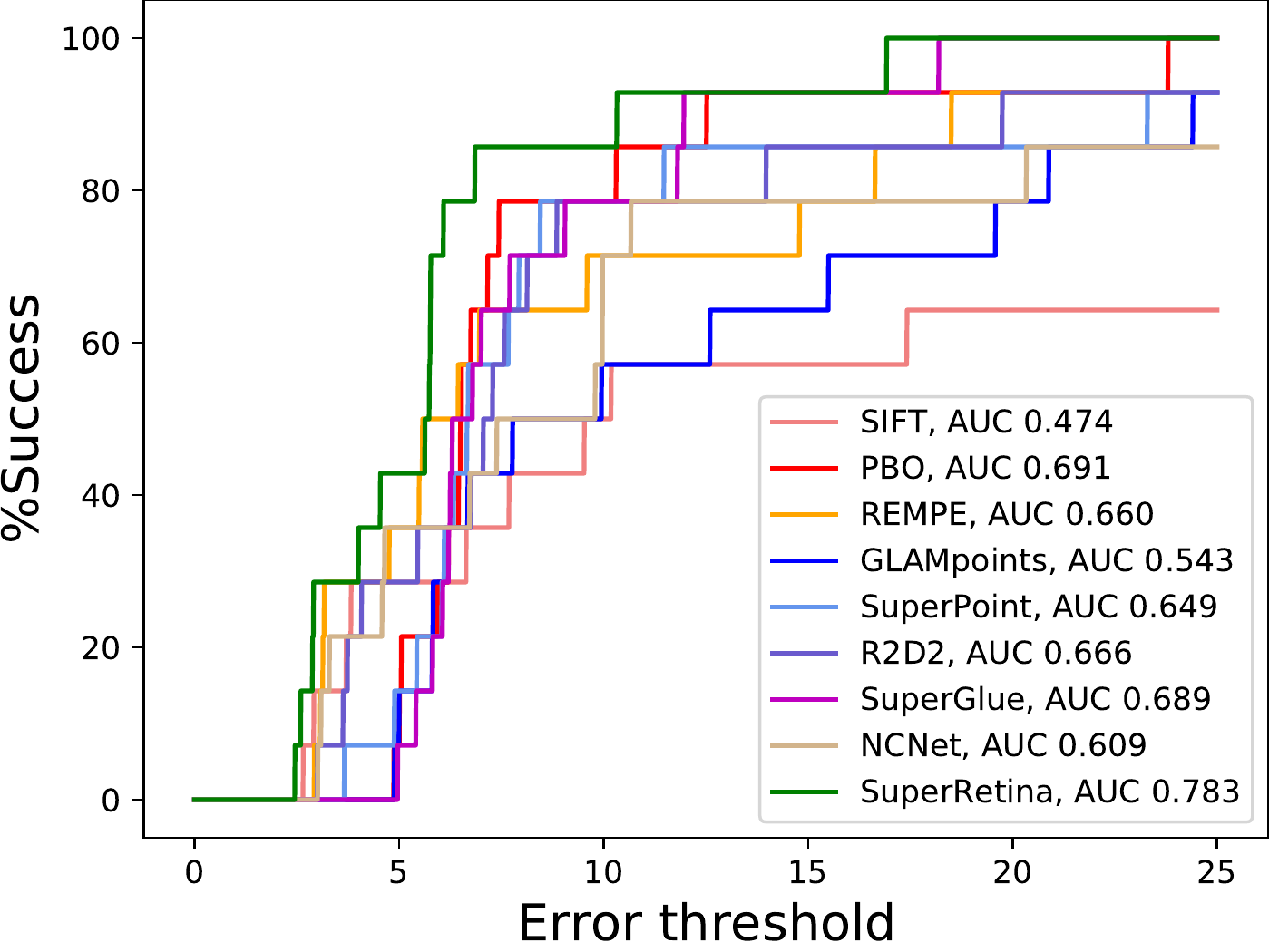}
         \caption{Moderate mode}
         \label{fig:auc-sota-Mod}
        \end{subfigure}
        \begin{subfigure}{.32\columnwidth}
         \includegraphics[width=\columnwidth]{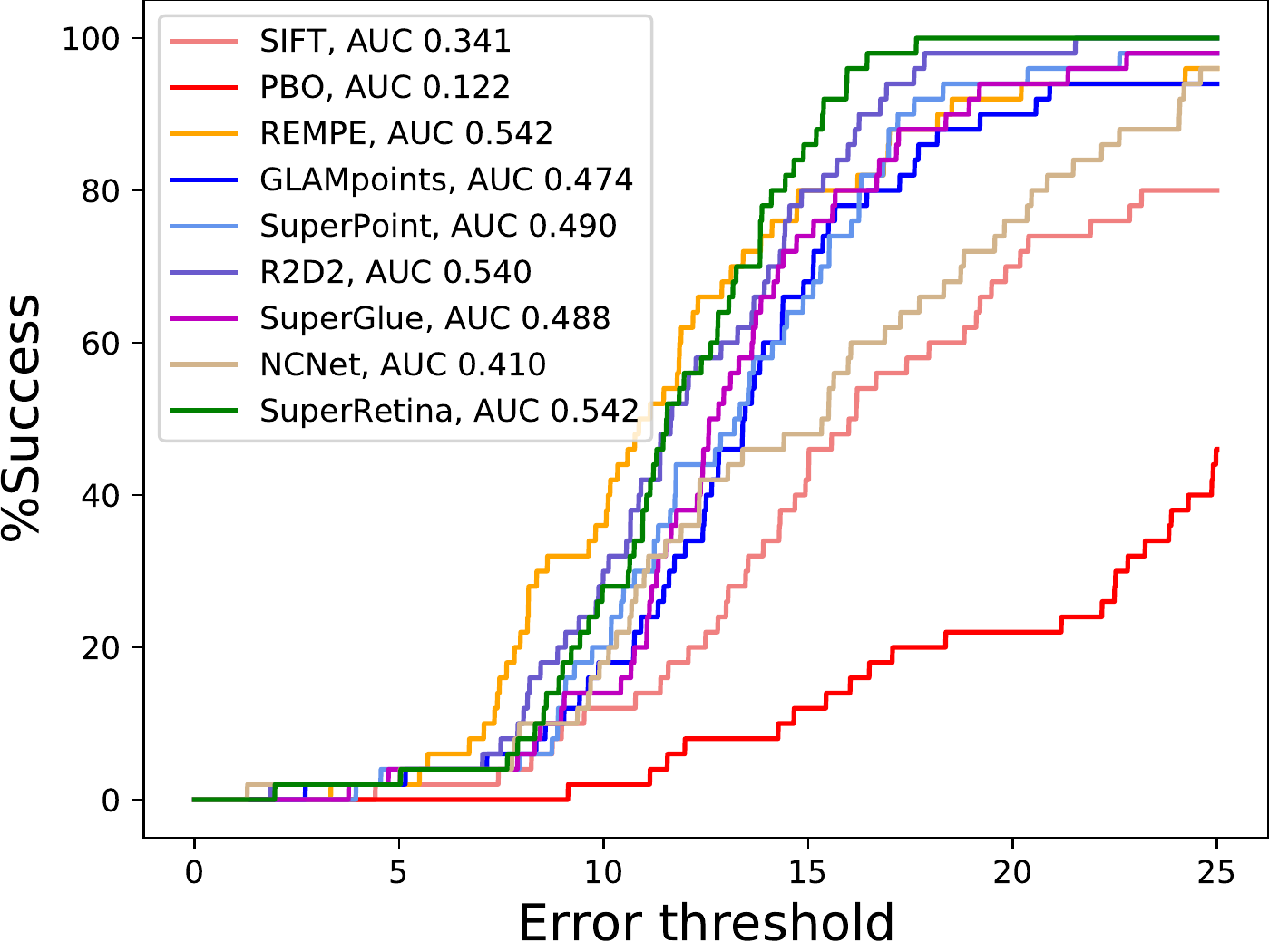}
         \caption{Hard mode}
         \label{fig:auc-sota-Hard}
         \end{subfigure}
      \caption{\textbf{Performance curves of the image registration successful rate \wrt the error threshold}. A curve  closer to the top left corner is better. The overall performance is measured by the Area Under the Curve (AUC) scores.} 
      \label{fig:auc}
   
\end{figure}

%% file: supplementary/fig-eer.tex
\begin{figure}[h!]
   \centering
   
        \begin{subfigure}{0.32\columnwidth}
         \includegraphics[width=\columnwidth]{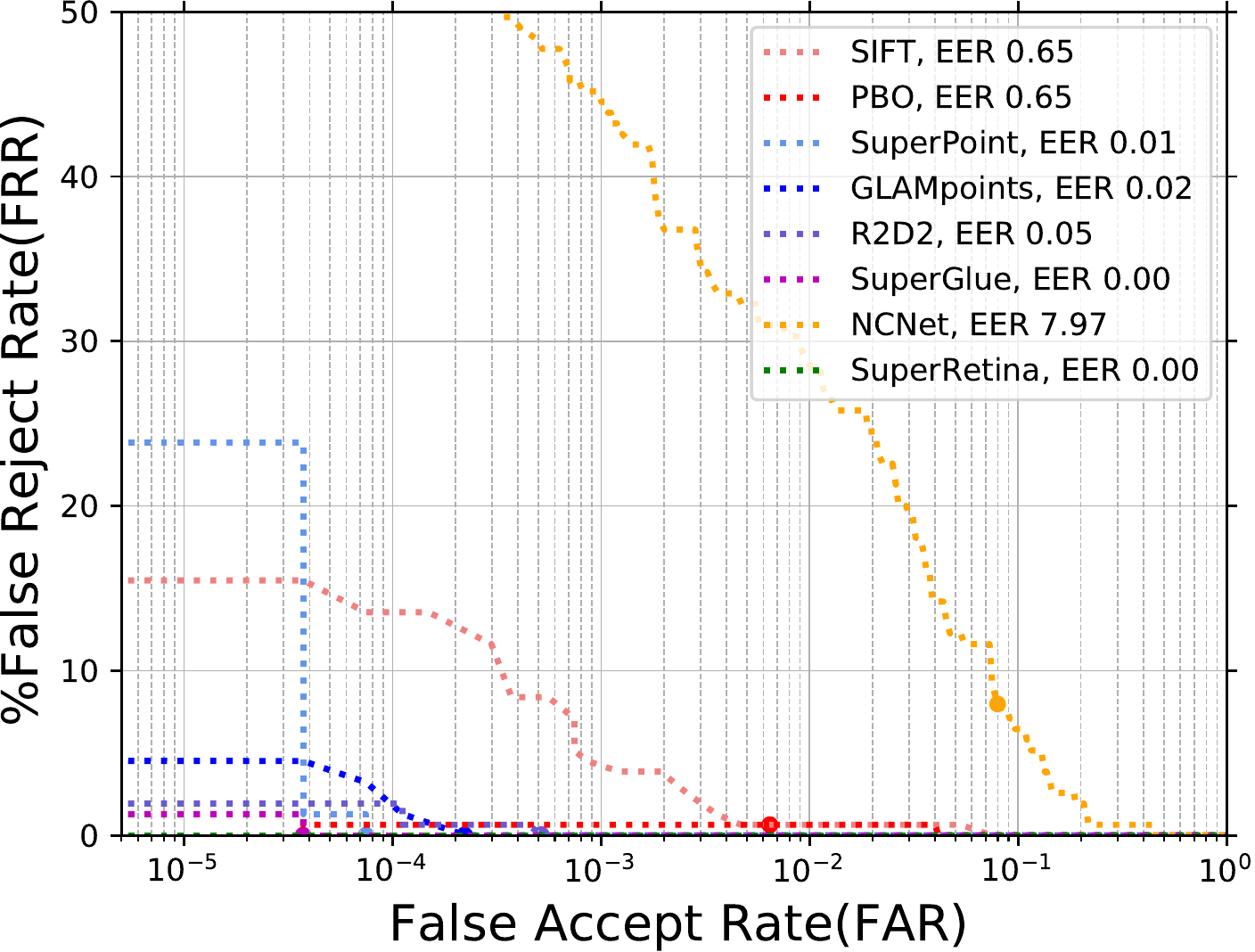}
         \caption{VARIA}
         \label{fig:eer-sota-VARIA}
         \end{subfigure}
         \begin{subfigure}{.32\columnwidth}
         \includegraphics[width=\columnwidth]{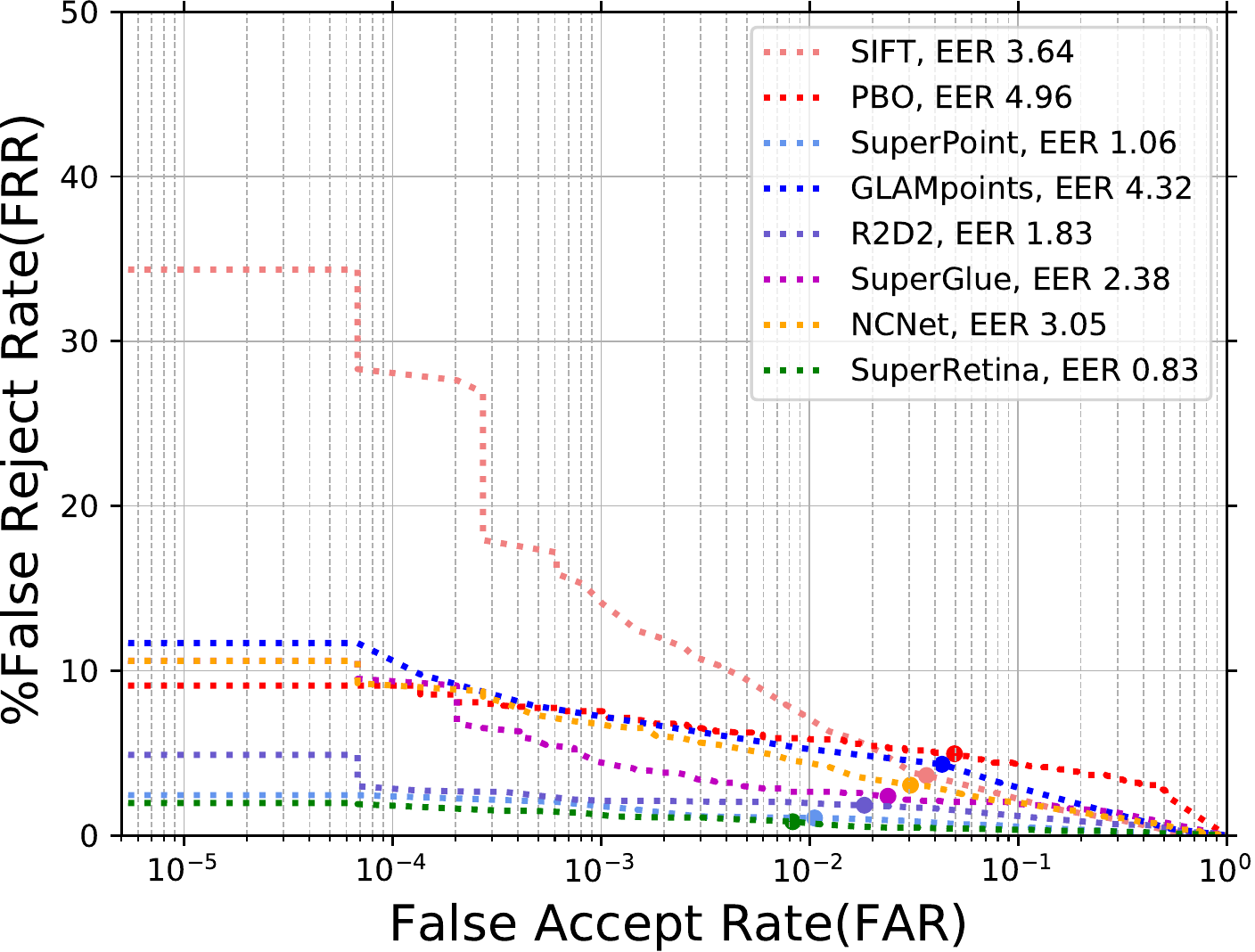}
         \caption{CLINICAL}
         \label{fig:eer-sota-CLINICAL}
        \end{subfigure}
        \begin{subfigure}{.32\columnwidth}
         \includegraphics[width=\columnwidth]{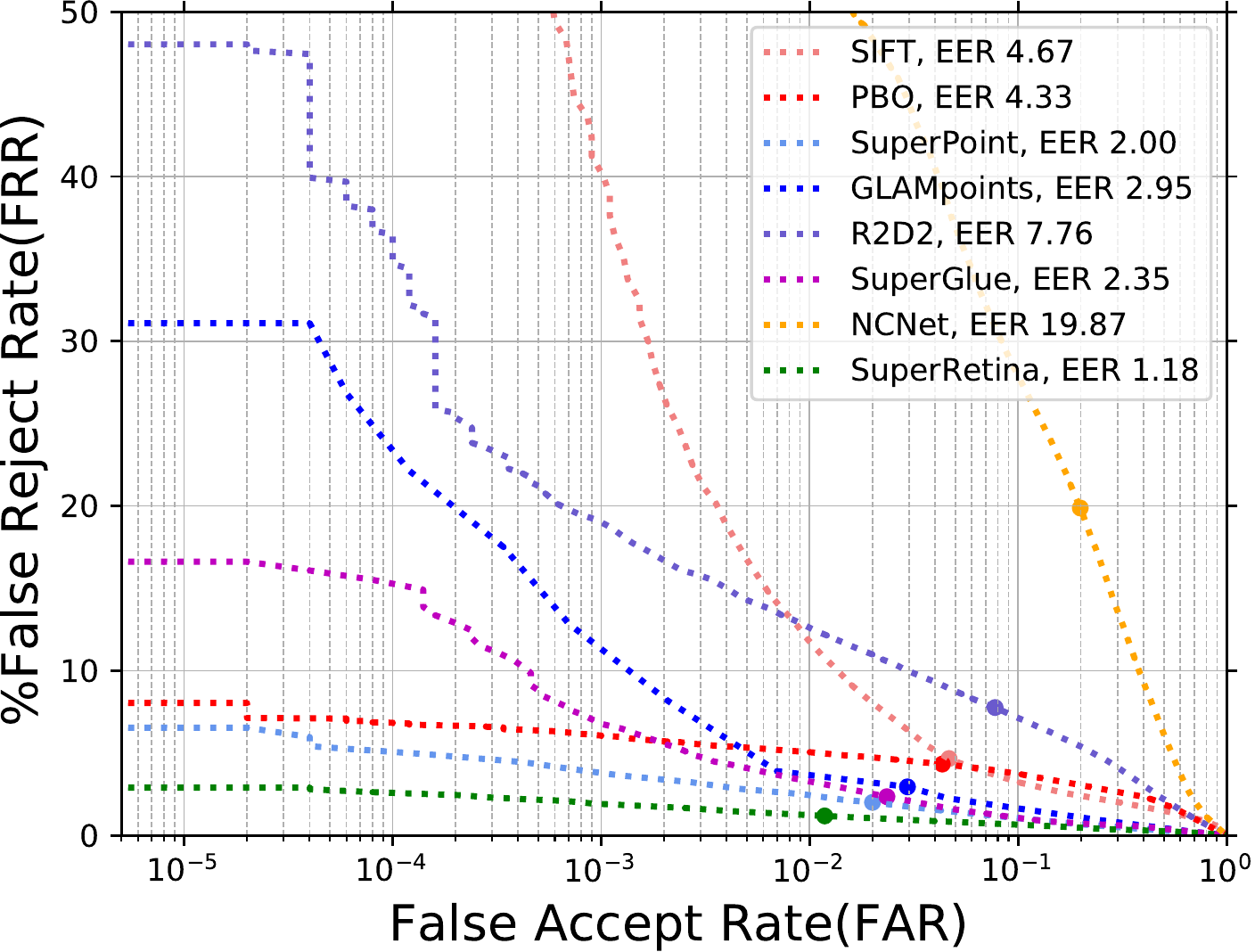}
         \caption{BES}
         \label{fig:eer-sota-BES}
         \end{subfigure}
         
      \caption{\textbf{Detection Error Tradeoff (DET) graphs on three test sets (a) VARIA, (b) CLINICAL and (c) BES for identity verification}. The equal error rate (EER) point per DET graph is shown in solid dots. Lower EER [\%] is better.}
        \label{fig:det}
\end{figure}

%% file: paper-5066.bbl
\begin{thebibliography}{10}
\providecommand{\url}[1]{\texttt{#1}}
\providecommand{\urlprefix}{URL }
\providecommand{\doi}[1]{https://doi.org/#1}

\bibitem{lospa-cvpr15}
Addison~Lee, J., Cheng, J., Hai~Lee, B., Ping~Ong, E., Xu, G., Wing Kee~Wong,
  D., Liu, J., Laude, A., Han~Lim, T.: A low-dimensional step pattern analysis
  algorithm with application to multimodal retinal image registration. In: CVPR
  (2015)

\bibitem{aleem2018fast}
Aleem, S., Sheng, B., Li, P., Yang, P., Feng, D.D.: Fast and accurate retinal
  identification system: Using retinal blood vasculature landmarks. IEEE
  Transactions on Industrial Informatics  \textbf{15}(7),  4099--4110 (2018)

\bibitem{arandjelovic2012three}
Arandjelovi{\'c}, R., Zisserman, A.: Three things everyone should know to
  improve object retrieval. In: CVPR (2012)

\bibitem{cattin2006retina}
Cattin, P.C., Bay, H., Van~Gool, L., Sz{\'e}kely, G.: Retina mosaicing using
  local features. In: MICCAI (2006)

\bibitem{chen2017deeplab}
Chen, L.C., Papandreou, G., Kokkinos, I., Murphy, K., Yuille, A.L.: {DeepLab}:
  Semantic image segmentation with deep convolutional nets, atrous convolution,
  and fully connected crfs. IEEE Transactions on Pattern Analysis and Machine
  Intelligence  \textbf{40}(4),  834--848 (2017)

\bibitem{dai2017scannet}
Dai, A., Chang, A.X., Savva, M., Halber, M., Funkhouser, T., Nie{\ss}ner, M.:
  {ScanNet}: Richly-annotated 3d reconstructions of indoor scenes. In: CVPR
  (2017)

\bibitem{detone2018superpoint}
DeTone, D., Malisiewicz, T., Rabinovich, A.: {SuperPoint}: Self-supervised
  interest point detection and description. In: CVPR Workshops (2018)

\bibitem{hernandez2020rempe}
Hernandez-Matas, C., Zabulis, X., Argyros, A.A.: {REMPE}: Registration of
  retinal images through eye modelling and pose estimation. IEEE Journal of
  Biomedical and Health Informatics  \textbf{24}(12),  3362--3373 (2020)

\bibitem{hernandez2017fire}
Hernandez-Matas, C., Zabulis, X., Triantafyllou, A., Anyfanti, P., Douma, S.,
  Argyros, A.A.: {FIRE}: Fundus image registration dataset. Modeling and
  Artificial Intelligence in Ophthalmology  \textbf{1}(4),  16--28 (2017)

\bibitem{jiang2021cotr}
Jiang, W., Trulls, E., Hosang, J., Tagliasacchi, A., Yi, K.M.: {COTR}:
  Correspondence transformer for matching across images. In: ICCV (2021)

\bibitem{BES1}
Jonas, J.B., Xu, L., Wang, Y.: The {B}eijing eye study. Acta Ophthalmologica
  \textbf{87}(3),  247--261 (2009)

\bibitem{kingma2014adam}
Kingma, D.P., Ba, J.: Adam: {A} method for stochastic optimization. In: ICLR
  (2015)

\bibitem{M2U-Net}
Laibacher, T., Weyde, T., Jalali, S.: {M2U-Net}: Effective and efficient
  retinal vessel segmentation for real-world applications. In: CVPRW (2019)

\bibitem{lajevardi2013retina}
Lajevardi, S.M., Arakala, A., Davis, S.A., Horadam, K.J.: Retina verification
  system based on biometric graph matching. IEEE Transactions on Image
  Processing  \textbf{22}(9),  3625--3635 (2013)

\bibitem{deepspa-iccv19}
Lee, J.A., Liu, P., Cheng, J., Fu, H.: A deep step pattern representation for
  multimodal retinal image registration. In: ICCV (2019)

\bibitem{lin2014microsoft}
Lin, T.Y., Maire, M., Belongie, S., Hays, J., Perona, P., Ramanan, D.,
  Doll{\'a}r, P., Zitnick, C.L.: Microsoft {COCO}: Common objects in context.
  In: ECCV (2014)

\bibitem{lowe2004distinctive}
Lowe, D.G.: Distinctive image features from scale-invariant keypoints.
  International Journal of Computer Vision  \textbf{60}(2),  91--110 (2004)

\bibitem{milletari2016v}
Milletari, F., Navab, N., Ahmadi, S.A.: {V-Net}: Fully convolutional neural
  networks for volumetric medical image segmentation. In: 3DV (2016)

\bibitem{oinonen2010identity}
Oinonen, H., Forsvik, H., Ruusuvuori, P., Yli-Harja, O., Voipio, V., Huttunen,
  H.: Identity verification based on vessel matching from fundus images. In:
  ICIP (2010)

\bibitem{ortega2009retinal}
Ortega, M., Penedo, M.G., Rouco, J., Barreira, N., Carreira, M.J.: Retinal
  verification using a feature points-based biometric pattern. EURASIP Journal
  on Advances in Signal Processing (235746) (2009)

\bibitem{r2d2}
Revaud, J., Weinzaepfel, P., de~Souza, C.R., Humenberger, M.: {R2D2}:
  Repeatable and reliable detector and descriptor. In: NeurIPS (2019)

\bibitem{rocco2018neighbourhood}
Rocco, I., Cimpoi, M., Arandjelovi{\'c}, R., Torii, A., Pajdla, T., Sivic, J.:
  Neighbourhood consensus networks. In: NeurIPS (2018)

\bibitem{ncnet}
Rocco, I., Cimpoi, M., Arandjelović, R., Torii, A., Pajdla, T., Sivic, J.:
  {NCNet}: Neighbourhood consensus networks for estimating image
  correspondences. IEEE Transactions on Pattern Analysis and Machine
  Intelligence  \textbf{44}(2),  1020--1034 (2022)

\bibitem{ronneberger2015u}
Ronneberger, O., Fischer, P., Brox, T.: {U-Net}: Convolutional networks for
  biomedical image segmentation. In: MICCAI (2015)

\bibitem{sarlin2020superglue}
Sarlin, P.E., DeTone, D., Malisiewicz, T., Rabinovich, A.: {SuperGlue}:
  Learning feature matching with graph neural networks. In: CVPR (2020)

\bibitem{sattler2018benchmarking}
Sattler, T., Maddern, W., Toft, C., Torii, A., Hammarstrand, L., Stenborg, E.,
  Safari, D., Okutomi, M., Pollefeys, M., Sivic, J., et~al.: Benchmarking
  {6DOF} outdoor visual localization in changing conditions. In: CVPR (2018)

\bibitem{schroff2015facenet}
Schroff, F., Kalenichenko, D., Philbin, J.: {FaceNet}: A unified embedding for
  face recognition and clustering. In: CVPR (2015)

\bibitem{simon1935new}
Simon, C.: A new scientific method of identification. New York state journal of
  medicine  \textbf{35}(18),  901--906 (1935)

\bibitem{sun2021loftr}
Sun, J., Shen, Z., Wang, Y., Bao, H., Zhou, X.: {LoFTR}: Detector-free local
  feature matching with transformers. In: CVPR (2021)

\bibitem{tian2019sosnet}
Tian, Y., Yu, X., Fan, B., Wu, F., Heijnen, H., Balntas, V.: {SOSNet}: Second
  order similarity regularization for local descriptor learning. In: CVPR
  (2019)

\bibitem{truong2019glampoints}
Truong, P., Apostolopoulos, S., Mosinska, A., Stucky, S., Ciller, C., Zanet,
  S.D.: {GLAMpoints}: Greedily learned accurate match points. In: ICCV (2019)

\bibitem{truong2021learning}
Truong, P., Danelljan, M., Van~Gool, L., Timofte, R.: Learning accurate dense
  correspondences and when to trust them. In: CVPR (2021)

\bibitem{wang2020segmentation}
Wang, Y., Zhang, J., An, C., Cavichini, M., Jhingan, M., Amador-Patarroyo,
  M.J., Long, C.P., Bartsch, D.U.G., Freeman, W.R., Nguyen, T.Q.: A
  segmentation based robust deep learning framework for multimodal retinal
  image registration. In: ICASSP (2020)

\bibitem{icpr20-lesion-net}
Wei, Q., Li, X., Yu, W., Zhang, X., Zhang, Y., Hu, B., Mo, B., Gong, D., Chen,
  N., Ding, D., Chen, Y.: Learn to segment retinal lesions and beyond. In: ICPR
  (2020)

\bibitem{wei2016convolutional}
Wei, S.E., Ramakrishna, V., Kanade, T., Sheikh, Y.: Convolutional pose
  machines. In: CVPR (2016)

\bibitem{BES2}
Wei, W., Xu, L., Jonas, J.B., Shao, L., Du, K., Wang, S., Chen, C., Xu, J.,
  Wang, Y., Zhou, J., You, Q.: Subfoveal choroidal thickness: The {B}eijing eye
  study. Ophthalmology  \textbf{120}(1),  175--180 (2013)

\end{thebibliography}
